\documentclass{article}

\usepackage[preprint]{corl_2026} %

\usepackage[numbers]{natbib}
\usepackage{multicol}

\usepackage{graphicx} 
\usepackage{wrapfig}

\usepackage{enumitem}

\usepackage[export]{adjustbox}
\usepackage{subcaption}

\usepackage{threeparttable}
\usepackage{makecell}
\usepackage{multirow}
\usepackage{booktabs} 

\usepackage{pifont}

\usepackage{algorithm}
\usepackage{algpseudocode}

\usepackage{amsmath}
\usepackage{amssymb}
\usepackage{amsfonts}

\usepackage{url}

\usepackage{afterpage}
\usepackage{placeins}
\newcommand{\mymethod}{SCOPE}
\title{Control-Ready Uncertainty for Trajectory Diffusion}

\author{
{\bfseries
Zhiwei Xue$^{1,*,\dagger}$, \quad
Jia Yue Kam$^{1,*}$, \quad
Jinhang Qiu$^{1,*}$, \quad
Yifeng Cheng$^{1}$,} \\
{\bfseries
Ege Gursoy$^{3}$, \quad
Jiaming Wang$^{1}$, \quad
Vincent Bonnet$^{3}$, \quad
and Harold Soh$^{1,2,\dagger}$} \\
$^{1}$National University of Singapore. 
$^{2}$Smart Systems Institute, NUS.
$^{3}$LAAS-CNRS. \\
$^{*}$Equal contribution. $^{\dagger}$Corresponding authors
}

\hypersetup{pdftitle={Control-Ready Uncertainty for Trajectory Diffusion},pdfauthor={},pdfsubject={},pdfkeywords={}}

\begin{document}
\maketitle

\begin{abstract}
Diffusion models can represent complex, multimodal trajectory distributions, but extracting uncertainty from them typically requires costly Monte Carlo sampling. This limits their use in real-time control, where robots must rapidly assess risk and maintain safety margins. We introduce \emph{Score-Curvature for Online Precision Estimation} (SCOPE), a lightweight module that augments diffusion trajectory models with control-ready uncertainty. SCOPE learns a structured precision matrix around each nominal trajectory by distilling score-curvature information and producing calibrated Gaussian tubes with low overhead and without repeated Monte Carlo sampling. These tubes provide per-timestep covariance estimates that can be used both as predicted occupancy for moving agents and as adaptive exploration guides for robot control. We evaluate SCOPE with mode-conditioned multimodal diffusion backbones in pedestrian forecasting, crowd navigation, Maze2D control, and real-world Franka Panda manipulation. Across these settings, SCOPE provides fast uncertainty estimation, which leads to better closed-loop performance. \href{https://zackaxue.github.io/SCOPE-project-page/}{Project page}.
\end{abstract}

\section{Introduction}
\label{Sec: Intro}
Diffusion models have recently gained wide adoption for their ability to represent complex and high-dimensional distributions. In robotics, they serve two complementary roles. As \emph{policies}, diffusion models generate robot motion plans or action sequences, enabling expressive behavior synthesis that captures the multimodality inherent in many manipulation and locomotion tasks~\citep{janner2022planning, chi2024diffusionpolicyvisuomotorpolicy}. 
As \emph{world models}, diffusion models can represent stochastic future evolutions of the environment, including scene dynamics, object motion, and interactions among agents. One important instance of this broader use case is trajectory prediction, where diffusion-based models~\citep{gu2022stochastictrajectorypredictionmotion, jiang2023motiondiffuser, mao2023leapfrogdiffusionmodelstochastic, bae2024singulartrajectoryuniversaltrajectorypredictor, zhong2022guidedconditionaldiffusioncontrollable} forecast the future motion of surrounding agents such as pedestrians, vehicles, or human collaborators; these predictions can then be used by controllers such as MPPI~\citep{mohamed2025c2umppi, trevisan2025dynamicriskawaremppimobile} to anticipate and avoid collisions.

However, in interactive robotics, accurate prediction is often insufficient. The robot should also be aware of the \emph{aleatoric uncertainty} in those predictions: the uncertainty arising from inherently stochastic future behavior. For example, a robot approaching a pedestrian must reason about both the pedestrian’s likely future trajectory and the ``spread'' around each plausible future before deciding whether to pass, yield, or maintain a safe distance. This control-relevant uncertainty allows the robot to assess collision risk and preserve safety margins. 

The challenge is to extract this uncertainty fast enough for control. Diffusion models represent uncertainty \emph{implicitly} through their generative distribution, and making it \emph{explicit} usually requires post hoc Monte Carlo sampling~\citep{neumeier2026uncertaintyawarediffusionmodelmultimodal, wintel2025usingdiffusionensemblesestimate}. Since each sample requires a reverse diffusion chain, reliable estimates can be too slow for closed-loop control even with GPU batching. This delay can leave the robot acting on an outdated motion forecast, increasing collision risk. Reducing the sample count can recover speed but degrades estimation fidelity. Hence, this raises a natural question: \textit{can we obtain control-ready uncertainty rapidly, without Monte Carlo sampling?}

\begin{wrapfigure}[28]{r}{0.36\textwidth}
    \centering
    \vspace{-6pt}
    \includegraphics[width=0.36\textwidth]{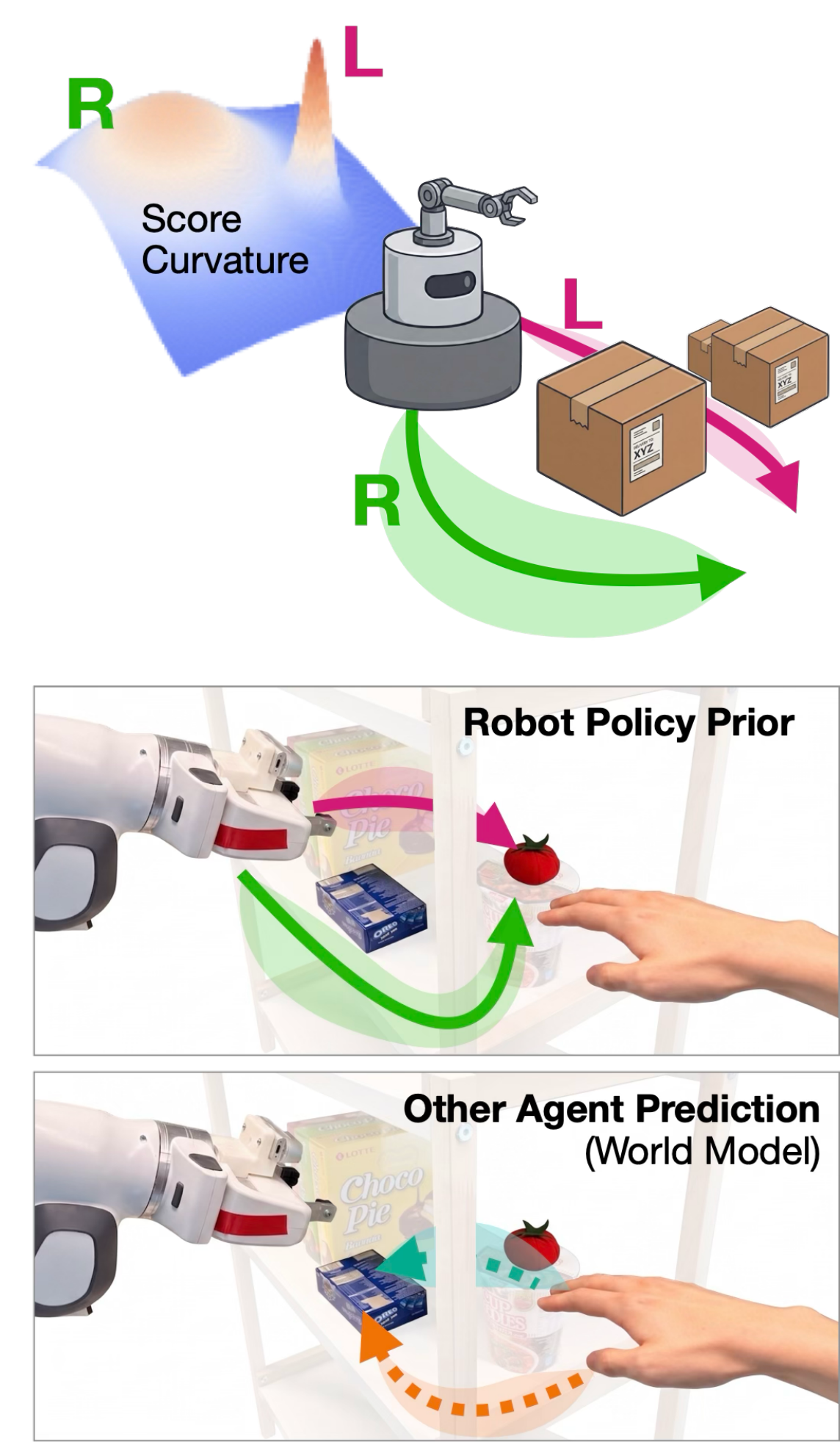}
    \caption{\small (\textbf{Top}) \mymethod{} leverages score curvature to augment diffusion predictions with precision tubes that are suitable for control. It provides uncertainty estimates at low computation cost, enabling real-time control. (\textbf{Bottom}) We apply \mymethod{} as a robot policy prior in MPPI and also to predict the trajectories of other agents.}
    \label{fig:teaser}
\end{wrapfigure}
In response, we propose \emph{Score Curvature for Online Precision Estimation} (\mymethod{}), a lightweight module that efficiently distills control-ready aleatoric trajectory uncertainty from a diffusion model. 
Our key observation is that a trained score model encodes the local geometry of the learned trajectory density (Fig.~\ref{fig:teaser}, top). 
Around a predicted trajectory, uncertainty depends on how quickly likelihood falls away in nearby directions: sharp decay yields a narrow, high-precision tube, while flat decay yields a wider region of plausible futures. 
Through Laplace's method, this local geometry corresponds to the curvature of the log-density, whose negative Hessian defines a precision matrix. But forming and querying this full (often large) curvature matrix at control time is impractical. Instead, \mymethod{} learns a structured precision head from score-curvature information using efficient Hessian-vector products (HVPs) during training. At inference, one diffusion pass followed by one lightweight head evaluation efficiently produces per-timestep marginal covariances, yielding a \textit{mixture of tubes}: multiple trajectory hypotheses, each equipped with analytic confidence regions that plug directly into controllers.

We evaluate \mymethod{} along three axes: uncertainty quality, inference speed, and downstream control performance. On pedestrian trajectory forecasting, \mymethod{} matches the uncertainty quality of diffusion ensembles while avoiding their costly repeated sampling; this makes uncertainty estimation fast enough for closed-loop control. We then demonstrate two control use-cases. As a policy-side guide, it uses trajectory tubes to adapt MPPI exploration in concave mazes, improving both sample efficiency and success rate. As a world model, \mymethod{} provides pedestrian motion tubes that allow MPPI to anticipate collision risk in crowded navigation. Finally, we validate practical viability on a real Franka Panda arm in a cluttered tabletop pick-and-place task (Fig.~\ref{fig:teaser}, bottom), which demonstrates closed-loop collision avoidance under tight control-rate constraints. 

To summarize, we make the following contributions: 
(i) We introduce \mymethod{}, a lightweight module that extracts control-ready, per-timestep trajectory uncertainty from a diffusion model with low overhead, avoiding repeated Monte Carlo sampling; 
(ii) We improve inference efficiency by parameterizing uncertainty as a block-banded plus low-rank precision matrix, which enables efficient HVP-based training and marginal covariance queries at test time; 
(iii) We validate \mymethod{} in forecasting, pedestrian navigation, Maze2D control, and real-robot manipulation, where it improves closed-loop performance over strong baselines.
\newcommand{\R}{\mathbb{R}}
\newcommand{\x}{\mathbf{x}}
\newcommand{\y}{\mathbf{y}}
\newcommand{\p}{\mathbf{p}}
\newcommand{\f}{\mathbf{f}}
\newcommand{\boldzero}{\mathbf{0}}
\newcommand{\boldepsilon}{\boldsymbol{\epsilon}}
\newcommand{\boldmu}{\boldsymbol{\mu}}
\newcommand{\boldSigma}{\boldsymbol{\Sigma}}
\newcommand{\calW}{\mathcal{W}}
\newcommand{\calA}{\mathcal{A}}
\newcommand{\calN}{\mathcal{N}}
\renewcommand{\u}{\mathbf{u}}
\renewcommand{\v}{\mathbf{v}}

\vspace{-0.3em}
\section{Preliminaries}\label{Sec: Prelim}
\vspace{-0.1em}
\textbf{Diffusion models as trajectory forecasters.}\label{sec:prelim_diffusion}
Diffusion models generate trajectories by learning to reverse a gradual noising process.
The forward process progressively corrupts a clean trajectory $\mathbf{y}_0$ into Gaussian noise over $\mathsf S$ steps, with closed form
$q(\mathbf{y}_{\mathsf s}|\mathbf{y}_0) = \mathcal{N}\!\big(\sqrt{\bar\alpha_{\mathsf s}}\,\mathbf{y}_0,\, (1-\bar\alpha_{\mathsf s})\mathbf{I}\big)$,
where $\bar\alpha_{\mathsf s}=\prod_{j=1}^{\mathsf s} \alpha_j$ and $\alpha_j\in(0,1)$ is a noise schedule.
A neural network then learns the reverse: given a noisy trajectory $\mathbf{y}_{\mathsf s}$ and context $\mathbf{c}$, it predicts a slightly less noisy $\mathbf{y}_{\mathsf s-1}$ via
$p_\theta(\mathbf{y}_{\mathsf s-1}|\mathbf{y}_{\mathsf s}, \mathbf{c})=\mathcal{N}\!\big(\boldsymbol{\mu}_\theta(\mathbf{y}_{\mathsf s},{\mathsf s},\mathbf{c}),\,\boldsymbol{\Sigma}_{\mathsf s}\big)$.
Starting from pure noise $\mathbf{y}_{\mathsf S}\sim\mathcal{N}(\mathbf{0},\mathbf{I})$, iterating this reverse step produces a sample from the learned trajectory distribution $p(\mathbf{y}_{1:T}\mid \mathbf{c})$.
Equivalently, the denoiser can be viewed as learning the \emph{score function} $s_\theta(\mathbf{y},\mathsf s)=\nabla_{\mathbf{y}}\log p_{\mathsf s}(\mathbf{y})$, i.e.\ the gradient of the log-density at noise level $\mathsf s$. At the terminal step ($\mathsf s\!=\!0$), $s_\theta(\mathbf{y},0)$ approximates $\nabla_{\mathbf{y}}\log p(\mathbf{y})$, the gradient of the clean data log-density, which we leverage in Sec.~\ref{sec:method_precision_head} to extract uncertainty.

\textbf{Model Predictive Path Integral (MPPI) Control.}
MPPI~\citep{williams2017mppi} is a sampling-based controller for non-linear systems $\x_{t+1}=\f(\x_t, \u_t)$ with state $\x_t\in\mathbb{R}^{n_x}$ and control $\u_t\in\mathbb{R}^{n_u}$.
It maintains a nominal control sequence $\{\u_t\}_{t=0}^{H-1}$ over a planning horizon $H$ and, at each iteration, draws $N$ random perturbations $\delta\u_t^{(k)} \sim \mathcal{N}(\mathbf{0}, \mathbf{\Sigma^u})$.
Each perturbed sequence $\u_t^{(k)}=\u_t+\delta\u_t^{(k)}$ is rolled out through the dynamics to obtain a trajectory cost $J^{(k)} = \sum_{t=0}^{H-1} \ell(\x_t^{(k)}, \u_t^{(k)})$.
The key idea is that lower-cost rollouts receive exponentially higher weight: MPPI updates the nominal controls via $\u_t \leftarrow \u_t + \sum_{k=1}^N w_k\,\delta\u_t^{(k)}$ where $w_k \propto \exp\!\big({-J^{(k)}}/{\lambda}\big)$ and $\lambda>0$ is a temperature.
This makes MPPI a flexible, gradient-free planner, but its performance depends heavily on the exploration covariance $\mathbf{\Sigma^u}$, which is a key limitation we address in this work.

\begin{figure*}[t]
    \centering
    \includegraphics[width=1\linewidth]{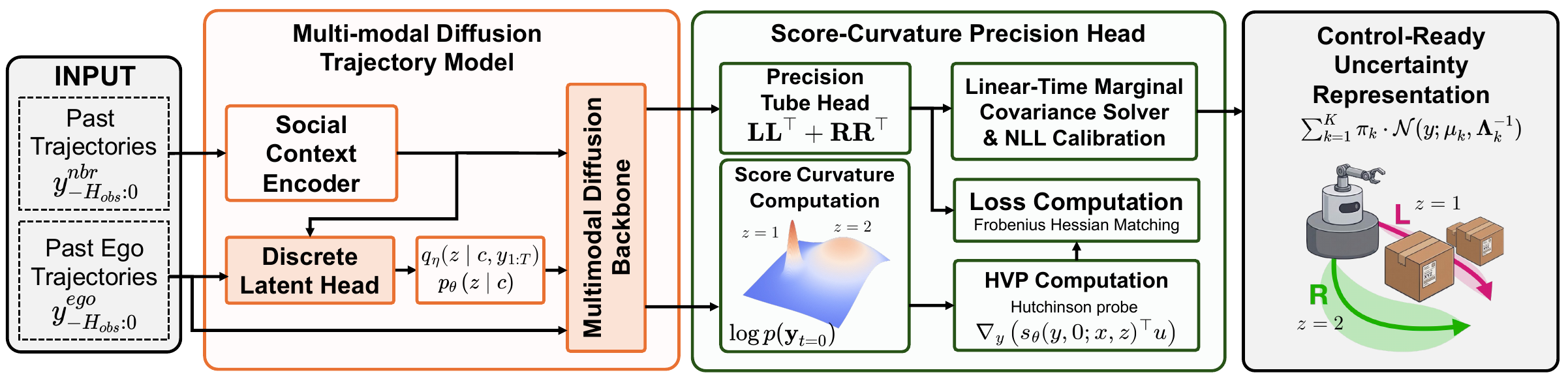}
    \caption{\small \textbf{\mymethod{} overview.} Our method extracts control-ready Gaussian tubes from a diffusion predictor. Here, we use a multimodal diffusion backbone (left) that produces $K$ nominal trajectories with explicit mode probabilities. The \mymethod{} Precision Head (right) wraps each trajectory in a structured precision matrix $\boldsymbol\Lambda_k$, learned by distilling the score model's terminal curvature via Hessian-vector products at training time. At inference, Gaussian tubes are produced with low overhead after diffusion decoding.}
    \label{Fig: pipeline}
    \vspace{-1.3em}
\end{figure*}

\section{Score-Curvature for Online Precision Estimation (SCOPE)}
\label{sec:method}

This section presents \mymethod{}, a lightweight module for extracting local, control-ready uncertainty from diffusion trajectory models. Given a conditioning context $\mathbf c$---such as agent histories, scene geometry, start and goal states, or task information---a diffusion model can produce one or more nominal future trajectories. \mymethod{} augments each nominal trajectory with a structured precision matrix, whose inverse defines a local Gaussian tube around that prediction. %

SCOPE can be applied to a single trajectory from a standard diffusion model, to multiple sampled trajectories, or to explicit modes produced by a multimodal diffusion backbone. Its local Laplace interpretation is best justified when the decoded trajectory lies near a local mode. In this paper, we instantiate SCOPE with the last option. As shown in Fig.~\ref{Fig: pipeline}, a multimodal diffusion backbone (Sec.~\ref{sec:method_mixture}) generates a small set of nominal trajectory hypotheses with explicit mode probabilities. SCOPE then predicts structured local uncertainty around each hypothesis using score-curvature information (Sec.~\ref{sec:method_precision_head}). The resulting representation is a mixture of Gaussian trajectory tubes that supports efficient marginal covariance queries and can be used both for predictive world modeling and for policy-side control guidance.

\subsection{Control Interface Representation: From Local Tubes to Mixtures}
\label{sec:gaussian_tube_interface}

For a single nominal trajectory $\boldsymbol\mu(\mathbf c)$, SCOPE predicts a structured precision matrix $\boldsymbol\Lambda(\mathbf c)\succ 0$. This defines a local Gaussian tube,
$
\mathcal{N}\left(
\mathbf y_{1:T};
\boldsymbol\mu(\mathbf c),
\boldsymbol\Lambda(\mathbf c)^{-1}
\right),
$
where $\boldsymbol\Lambda(\mathbf c)^{-1}$ captures local trajectory spread around the nominal prediction.
When the backbone provides $K$ nominal hypotheses with mode probabilities, these local tubes combine into a mixture representation:
\begin{equation}
p(\mathbf y_{1:T}\mid \mathbf c)
\;\approx\;
\sum\nolimits_{k=1}^{K} \pi_k(\mathbf c)\,
\mathcal{N}\!\big(
\mathbf y_{1:T};
\boldsymbol\mu_k(\mathbf c),
\boldsymbol\Lambda_k(\mathbf c)^{-1}
\big),
\label{eq:gaussian_tube_output}
\end{equation}
where $\boldsymbol\mu_k(\mathbf c)\in\mathbb{R}^{dT}$ is the nominal trajectory for mode $k$, $\pi_k(\mathbf c)$ is its mode probability with $\sum_k \pi_k(\mathbf c)=1$, and $\boldsymbol\Lambda_k(\mathbf c)\succ 0$ is a structured joint precision matrix over the full trajectory. The marginal covariance blocks of $\boldsymbol\Lambda_k(\mathbf c)^{-1}$ provide the per-timestep uncertainty.
We call each mixture component a \emph{Gaussian tube}. The mode probability $\pi_k$ captures uncertainty over distinct futures, while the inverse precision $\boldsymbol\Lambda_k^{-1}$ captures local trajectory spread within a particular mode. Thus, the multimodal backbone provides a set of hypotheses, and SCOPE turns each hypothesis into an analytic uncertainty tube suitable for control.

\subsection{Multimodal Diffusion Backbone}
\label{sec:method_mixture}

A standard diffusion model can produce diverse futures by sampling, but the resulting samples do not come with explicit mode identities or probabilities and typically require post-hoc clustering before they can be used as a control interface. This is undesirable for real-time control, where the planner should evaluate risk over a small number of interpretable hypotheses rather than over a large cloud of samples.

We therefore instantiate SCOPE with a discrete-latent multimodal diffusion backbone. This choice separates two kinds of uncertainty: the discrete latent variable captures global ambiguity over qualitatively different futures, while SCOPE estimates local trajectory spread around each future. This separation is also important geometrically. When plausible futures correspond to distinct intentions or routes, a single continuous predictor can blur across incompatible modes and assign probability to unrealistic intermediate trajectories. Explicit modes avoid this averaging effect by giving each hypothesis its own local density model, making the subsequent Gaussian-tube approximation both more meaningful and useful for control.

We follow the mixture formulation of Trajectron++~\citep{salzmann2021trajectrondynamicallyfeasibletrajectoryforecasting} and model the predictive distribution as
\begin{align}
p_{\theta}(\mathbf y_{1:T}\mid \mathbf c)
=
\sum\nolimits_{k=1}^{K} p_{\theta}(z{=}k\mid \mathbf c)\,
p_{\psi}(\mathbf y_{1:T}\mid \mathbf h_\mathbf c, z{=}k),
\label{eq:mixture_diffusion}
\end{align}
where $z\in\{1,\dots,K\}$ indexes the mode, $\mathbf c$ is the observed context, and $\mathbf h_\mathbf c$ is the encoder feature. The categorical prior $p_{\theta}(z|\mathbf c)$ gives the mode probabilities, while the conditional diffusion decoder $p_{\psi}(\mathbf y_{1:T}|\mathbf h_\mathbf c,z)$ models the trajectory distribution within each mode.
As shown in Fig.~\ref{Fig: pipeline}~(left), a \emph{social context encoder} maps observed trajectories to $\mathbf h_\mathbf c$, and a \emph{discrete latent head} produces the mode prior $p_{\theta}(z|\mathbf c)$. For each mode $k$, the diffusion decoder produces a nominal trajectory $\boldsymbol\mu_k$ with probability $\pi_k := p_\theta(z{=}k|\mathbf c)$. SCOPE then wraps each nominal trajectory with a structured precision tube, yielding $K$ hypotheses without post-hoc sample clustering.

\noindent\textbf{Training objective.}
Following the InfoVAE~\citep{zhao2019infovae} formulation, we introduce an encoder $q_\eta(z|\mathbf c, \mathbf y)$ and maximize the ELBO, writing $\mathbf y$ for $\mathbf y_{1:T}$:
\begin{align}
\mathcal{L}_{\text{ELBO}} =
\mathbb{E}_{q_\eta(z|\mathbf c, \mathbf y)}
\!\big[\mathcal{L}_{\text{diff}}(\mathbf c, \mathbf y; z)\big]
-
\beta\,\mathrm{KL}\!\big(
q_\eta(z|\mathbf c, \mathbf y)
\,\|\,
p_\theta(z|\mathbf c)
\big),
\label{eq:elbo}
\end{align}
where $\mathcal{L}_{\text{diff}}$ is the DDPM variational bound of the $z$-conditioned decoder and $\beta$ is the KL weight. Details are provided in Appendix~\ref{app:mixture_training}.

\subsection{Score-Curvature Precision Head}
\label{sec:method_precision_head}
The diffusion backbone provides nominal trajectories and mode probabilities, but not a structured uncertainty estimate suitable for real-time control. Extracting uncertainty via Monte Carlo sampling, drawing hundreds of diffusion trajectories and fitting a Gaussian post-hoc, is too expensive at control time. \mymethod{} addresses this by predicting a structured local precision matrix around each nominal trajectory in a single forward pass.

\noindent\textbf{Local uncertainty from score curvature.}
For each mode $k$, the backbone produces a nominal trajectory $\boldsymbol\mu_k$ via deterministic DDIM decoding conditioned on the discrete latent $z{=}k$. 
We use the decoded trajectory $\boldsymbol\mu_k$ as a near-modal anchor $\hat{\mathbf y}$, supported by the small displacements from local quadratic centers observed in our empirical diagnostics (Appendix~\ref{app:anchor_diagnostics}). Our key observation is that the curvature of the log-density at $\hat{\mathbf y}$ encodes local uncertainty:
\begin{equation}
\boldsymbol\Lambda^{*}(\hat{\mathbf y}) \;:=\; -\nabla_\mathbf y^2 \log p(\mathbf y)\big|_{\mathbf y=\hat{\mathbf y}},
\label{eq:lambda_star_def}
\end{equation}
which serves as a local precision signal around $\hat{\mathbf y}$. Where the log-density has sharp curvature (a narrow peak), the model is confident that the trajectory passes close to $\hat{\mathbf y}$, yielding high precision; where the curvature is flat, nearby trajectories are almost equally likely, indicating large uncertainty. Under Laplace's method, a smooth log-density is locally quadratic near a mode, giving $p(\mathbf y\mid z\!=\!k, \mathbf c)\approx \mathcal{N}(\mathbf y;\,\hat{\mathbf y},\,(\boldsymbol\Lambda^{*}(\hat{\mathbf y}))^{-1})$ (Appendix~\ref{app:laplace_approximation}), which yields a per-mode uncertainty estimate from the learned density.

\noindent\textbf{Ideal learning objective.}
We train a lightweight neural network, parameterized by $\phi$, to approximate the teacher precision:
$\boldsymbol\Lambda_\phi(\hat{\mathbf y}, \mathbf c, z) \approx \boldsymbol\Lambda^{*}(\hat{\mathbf y})$.
The ideal training objective minimizes the Frobenius-norm discrepancy between the predicted and teacher precision:
\begin{equation}
\mathcal{L}_{\text{ideal}}(\phi)
\;=\;
\mathbb{E}_{\hat{\mathbf y}}\!\Big[\big\|\boldsymbol\Lambda_\phi(\hat{\mathbf y}, \mathbf c, z) - \boldsymbol\Lambda^{*}(\hat{\mathbf y})\big\|_F^2\Big].
\label{eq:ideal_objective}
\end{equation}
Two challenges remain: (i)~$\boldsymbol\Lambda_\phi$ must be structured for efficient control-time covariance queries, and (ii)~the dense teacher $\boldsymbol\Lambda^*$ is prohibitively expensive to materialize, so the objective above cannot be computed directly. We address these in turn.

\noindent\textbf{Structured precision for efficient queries.}
Directly predicting a dense precision matrix is expensive and does not support efficient control-time marginal queries. We therefore parameterize $\boldsymbol\Lambda_\phi$ as a structured SPD matrix, predicted by an MLP. Because nearby timesteps are strongly coupled through dynamics while distant ones are approximately conditionally independent given intermediate states, the precision matrix of a trajectory is approximately block banded (Appendix~\ref{app:markov_sparsity}). We exploit this by restricting $\boldsymbol\Lambda_\phi$ to block bandwidth $b$ with a low-rank term for long-range correlations:
\begin{equation}
\boldsymbol\Lambda_\phi(\mathbf y)=\mathbf A_\phi(\mathbf y) + \mathbf R_\phi(\mathbf y)\mathbf R_\phi(\mathbf y)^\top,
\quad
\mathbf A_\phi(\mathbf y)=\mathbf L_\phi(\mathbf y)\mathbf L_\phi(\mathbf y)^\top,
\label{eq:precision_param_main}
\end{equation}
where $\mathbf L_\phi(\mathbf y)$ is the block lower-banded Cholesky factor of $\mathbf A_\phi \succ 0$, and $\mathbf R_\phi(\mathbf y)\in\mathbb{R}^{dT\times r}$ ($r\ll dT$) is a low-rank correction (Appendix~\ref{app:chol_structure}). This structure enables fast marginal covariance queries via banded Cholesky solves, with the low-rank term handled by a Woodbury update (Appendix~\ref{app:inference_complexity}).

\noindent\textbf{Training with HVPs and Hutchinson probes.}
The teacher $\boldsymbol\Lambda^{*}_k$ is never formed explicitly. Instead, for any probe vector $\mathbf u$, its response to that probe can be computed through a Hessian-vector product:
$
\boldsymbol\Lambda^{*}_k(\mathbf y,\mathbf c)\mathbf u
=
-\nabla_\mathbf y
\left(
s_\theta(\mathbf y,0;\mathbf c,z{=}k)^\top \mathbf u
\right),
$
computed by automatic differentiation (Appendix~\ref{app:hvp_identity}). After freezing the diffusion backbone, the precision head is trained by drawing $Q$ Hutchinson probes $\mathbf u_q\sim\mathcal{N}(\mathbf 0,\mathbf I)$. By the Hutchinson trace identity (Appendix~\ref{app:hutchinson}), $\mathbb{E}_{\mathbf u}[\|(\boldsymbol\Lambda_\phi - \boldsymbol\Lambda^*)\mathbf u\|^2] = \|\boldsymbol\Lambda_\phi - \boldsymbol\Lambda^*\|_F^2$ for isotropic $\mathbf u$, so Eq.~\eqref{eq:ideal_objective} can be optimized through random probes. We therefore minimize:
\begin{equation}
\mathcal{L}_{\text{prec}}(\phi)
\;=\;
\mathbb{E}_{\mathbf y}\,\mathbb{E}_{\mathbf u}\Big[\,\big\|\boldsymbol\Lambda_\phi(\mathbf y)\mathbf u
\;+
\nabla_\mathbf y\big(s_\theta(\mathbf y,0)^\top \mathbf u\big)\big\|_2^2\,\Big],
\label{eq:prec_loss_hvp}
\end{equation}
which moves all second-order computation to training time. At inference, one diffusion pass produces the anchors, one SCOPE forward pass predicts structured precisions, and per-timestep marginal covariances are recovered with the structured solver. Finally, we apply a lightweight post-hoc calibration by optimizing per-horizon scaling factors for trajectory-level negative log-likelihood on a held-out set (Appendix~\ref{sec:nll_calibration}).

\subsection{Using SCOPE in Downstream Control}
\label{sec:downstream_control}

We use SCOPE's Gaussian tubes in two complementary control settings. First, when the diffusion model acts as a \emph{world model}, the tubes represent uncertainty over other agents' future motion. For agent $n$ at time $t$, SCOPE gives a Gaussian mixture over position,
$
\mathbf p^{(n)}_t
\sim
\sum_{k=1}^{K}
\pi^{(n)}_k
\mathcal{N}\!\left(
\boldsymbol\mu^{(n)}_{k,t},
\tilde{\boldsymbol\Sigma}^{(n)}_{k,t}
\right),
$
where $\tilde{\boldsymbol\Sigma}^{(n)}_{k,t}=\tau_t\boldsymbol\Sigma^{(n)}_{k,t}$ is the calibrated marginal covariance and $\tau_t>0$ is a scalar from the calibration result. MPPI uses this distribution to penalize candidate rollouts by their probability of collision with predicted agent occupancies; the chance-cost formulation is given in Appendix~\ref{app:mppi_cost}.

Second, when the diffusion model acts as a \emph{policy prior}, the tubes describe flexibility around the robot's own predicted plan. We use the diffusion trajectory to initialize MPPI's nominal control sequence and to provide subgoals for the running cost. SCOPE then adapts MPPI's exploration covariance over time:
$
\delta \mathbf u^{(k)}_t
\sim
\mathcal{N}(\mathbf 0,\boldsymbol\Sigma_{t}^{\mathbf u})$ and 
$\boldsymbol\Sigma_{t}^{\mathbf u}
=
\frac{\gamma^2}{\Delta t^2}
\boldsymbol\Sigma_{t+1}^{\mathbf x},
$
where $\boldsymbol\Sigma_{t+1}^{\mathbf x}$ is the tube covariance at the next waypoint and $\gamma<1$ is a tube-radius scale factor. Large covariance expands exploration in flexible regions, while small covariance contracts exploration in narrow or high-precision regions. 
This lets MPPI allocate samples according to local flexibility, rather than using a fixed proposal scale throughout the trajectory.

\begin{figure*}[t]
    \begin{minipage}[c]{0.4\textwidth}
        \centering
        \includegraphics[width=\textwidth]{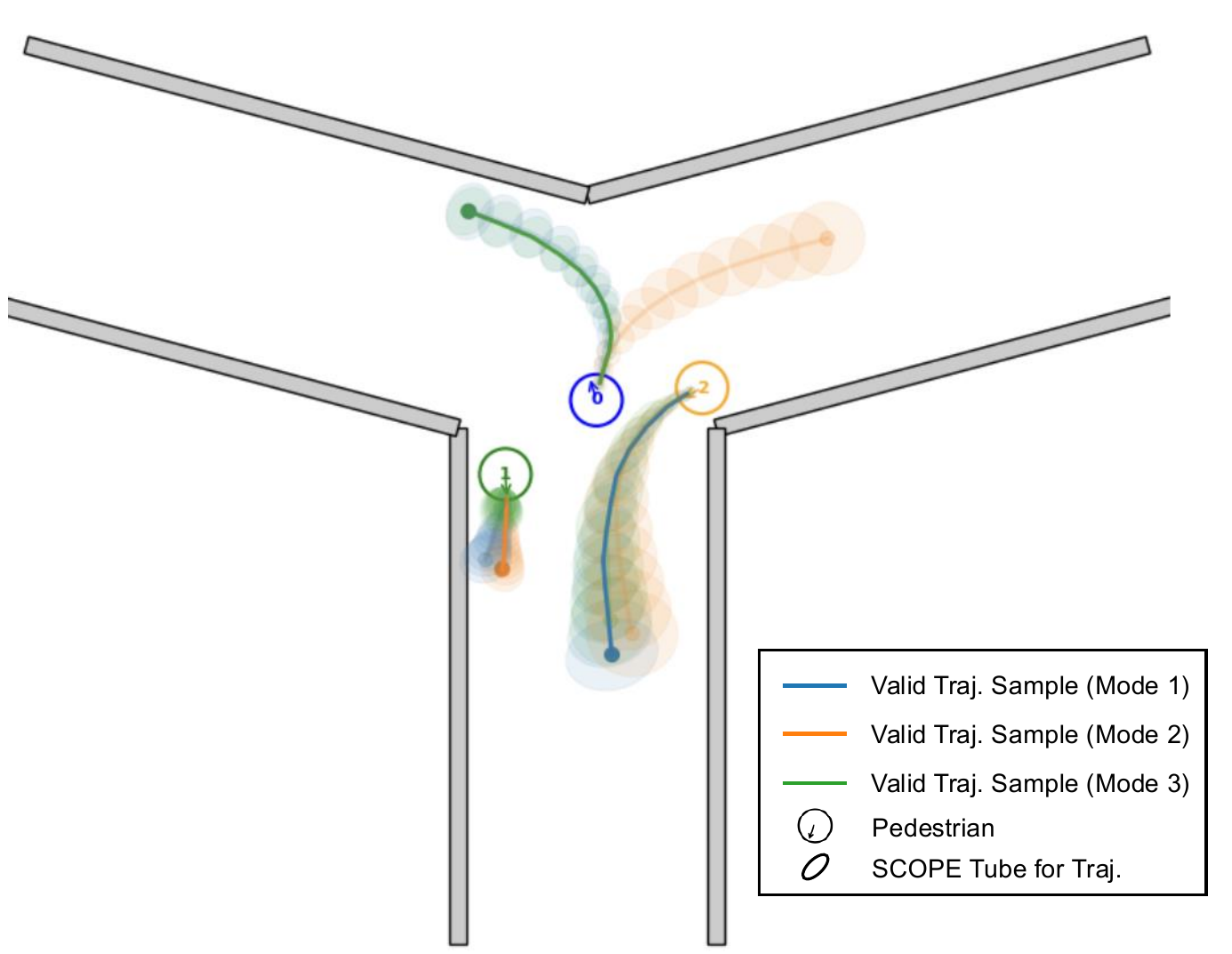}
    \end{minipage}
    \hfill
    \begin{minipage}[c]{0.55\textwidth}
        \centering
        \scriptsize
        \setlength{\tabcolsep}{3pt}
        \begin{threeparttable}
\begin{tabular}{l|cccc}
          \toprule
          & \multicolumn{4}{c}{\textbf{Negative Log-Likelihood (NLL) $\downarrow$}} \\
          \textbf{Method} & $N_p{=}3$ & $N_p{=}6$ & $N_p{=}9$ & $N_p{=}12$ \\
          \midrule
          LKF              & 9.9{\tiny$\pm$16.2} & 9.0{\tiny$\pm$17.0} & 6.8{\tiny$\pm$14.0} & 5.0{\tiny$\pm$12.6}
   \\
          MID-Ens-64       & -53.0{\tiny$\pm$5.3} & -31.0{\tiny$\pm$67.0} & -20.9{\tiny$\pm$86.3} &
  -20.1{\tiny$\pm$68.6} \\
          MID-Ens-128      & {-56.5}{\tiny$\pm$4.0} & -35.9{\tiny$\pm$62.2} & -40.5{\tiny$\pm$38.6} &
  -31.5{\tiny$\pm$60.5} \\
          MID-Ens-256      & \underline{-58.0}{\tiny$\pm$3.5} & -50.4{\tiny$\pm$19.8} & -50.8{\tiny$\pm$18.3} &
  -43.6{\tiny$\pm$35.3} \\
          \cmidrule{1-5}
          \textbf{\mymethod{} Variants:}\\
          Diag             & -39.3{\tiny$\pm$29.0} & -34.1{\tiny$\pm$35.7} & -32.6{\tiny$\pm$47.5} &
  -35.2{\tiny$\pm$41.0}\\
          Banded           & -52.8{\tiny$\pm$14.0} & -34.5{\tiny$\pm$33.4} & -39.9{\tiny$\pm$42.3} &
  -49.0{\tiny$\pm$21.9} \\
          Banded+LR (no cal.) & -57.5{\tiny$\pm$13.6} & \underline{-52.1}{\tiny$\pm$17.0} &
  \underline{-55.0}{\tiny$\pm$17.4} &
    \underline{-49.7}{\tiny$\pm$34.1} \\
          Banded+LR (Ours) & \textbf{-61.4}{\tiny$\pm$6.4} & \textbf{-52.2}{\tiny$\pm$12.2} &
  \textbf{-56.4}{\tiny$\pm$13.0} &
    \textbf{-60.6}{\tiny$\pm$11.0} \\
          \bottomrule
\end{tabular}
        \begin{tablenotes}[flushleft]\scriptsize
            \item \textbf{Bold}: best, \underline{underline}: second best.
        \end{tablenotes}
        \end{threeparttable}
    \end{minipage}
    \caption{(\textbf{Left}) Y-corridor environment. (\textbf{Right}) Uncertainty quality (NLL).}
    \label{fig:ph_nll_freq}
    \label{tab:ph_nll_freq}
    \label{fig:y_corridor}
    \vspace{-1em}
\end{figure*}

\section{Experiments}
\label{Sec: Experiments}
Our experiments are designed to answer three questions:
\textbf{(Q1)}~Does \mymethod{} produce accurate uncertainty tubes with fast inference, compared to sampling-based alternatives?
\textbf{(Q2)}~When used for policy, does \mymethod{} improve MPPI rollout efficiency and downstream control performance?
\textbf{(Q3)}~When used as a predictive world model, does \mymethod{} improve downstream control in interactive scenarios? 

\subsection{SCOPE Precision Head Properties}\label{exp:ph_property}

\begin{wraptable}{r}{0.38\textwidth}
    \centering
    \vspace{-12pt}
    \caption{Inference time breakdown of \mymethod{} (Banded+LR).}
    \label{tab:inference_time}
    \small
    \vspace{-5pt}
    \begin{tabular}{l|c}
        \toprule
        \textbf{Component} & \textbf{Time [ms]} \\
        \midrule
        Diffusion backbone & 12.19 \\
        SCOPE Precision Head     & 0.60 \\
        Calibration Scaling       & 0.08 \\
        \midrule
        \textbf{Total}     & \textbf{12.87} \\
        \bottomrule
    \end{tabular}
    \vspace{-10pt}
\end{wraptable}
We evaluate \mymethod{} on pedestrian trajectory forecasting in an ORCA-simulated Y-corridor environment~\citep{van2011reciprocal} (Fig.~\ref{fig:y_corridor}, left) with $N_p\in\{3,6,9,12\}$ pedestrians, reporting Negative Log-Likelihood (NLL) as the primary uncertainty metric.
We compare against \textbf{LKF}, a constant-velocity Kalman filter, and \textbf{MID-Ensemble}, which draws $M\in\{64,128,256\}$ diffusion samples and fits a GMM to the resulting trajectory set.
We ablate the precision structure in Eq.~\ref{eq:precision_param_main},
$\Lambda = LL^\top + RR^\top$: \textbf{Diag} uses independent per-timestep precision,
\textbf{Banded} keeps only the block-banded $LL^\top$ term, 
\textbf{Banded+LR (no cal.)} adds $RR^\top$ without calibration, and 
\textbf{Banded+LR} is the full calibrated model.
All variants differ only in how the precision tube head parameterizes uncertainty.

\textbf{Results.} The \mymethod{} precision head delivers high-quality uncertainty estimates with negligible overhead, answering Q1 in the affirmative. 
In uncertainty quality (Fig.~\ref{tab:ph_nll_freq}, right), \mymethod{} matches or improves over the strongest ensemble baselines across all crowd densities.
The ablations show that structure matters: banded precision improves over diagonal precision by modeling temporal correlations, while the low-rank term and calibration further improve trajectory-level NLL. Table~\ref{tab:inference_time} shows that the precision tube head and calibration take less than 1\,ms on an A5000, enabling real-time control. Even with fully batched GPU inference, MID-Ensemble diffusion cost grows with the sample budget, making it up to $7\times$ slower (Appendix~\ref{app:inference_time}).

\subsection{Case Study 1: Maze2D Navigation}\label{exp:maze2d}

To evaluate whether \mymethod{} improves MPPI rollout efficiency when used for policy (Q2), we first test \mymethod{} on a suite of 2D maze environments with randomly generated non-convex layouts; training and test mazes use different configurations.
Given a random start and goal position, the robot must navigate to its goal through the maze without collision, within a fixed timeout.
Pure diffusion planners~\citep{janner2022planning,luo2024potentialbaseddiffusionmotion} cannot guarantee constraint satisfaction in novel geometry, motivating closed-loop MPPI refinement whose efficiency depends on the exploration covariance.
We compare our method (Diff-MPPI + \mymethod{}) against three MPPI baselines:
\textbf{Vanilla MPPI}~\citep{williams2017mppi}, which samples perturbations from a fixed isotropic Gaussian;
\textbf{Log-MPPI}~\citep{mohamed2022autonomousnavigationagvsunknown}, which uses normal log-normal perturbations for efficient exploration with small noise variance;
and \textbf{Diff-MPPI}, which warm-starts MPPI with a diffusion-generated nominal trajectory but uses a fixed exploration covariance (i.e., diffusion-guided but without \mymethod{}). Extra comparisons against pure diffusion planners are in Appendix~\ref{app:maze2d_results}.

\begin{figure*}[t]
    \centering
    \includegraphics[width=\textwidth]{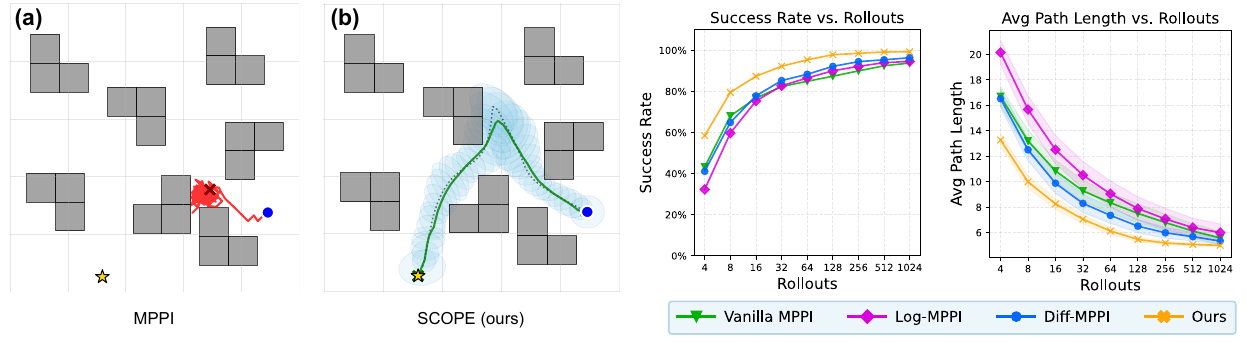}
    \caption{(\textbf{Left}) Maze2D. (\textbf{Right}) SR and average path length vs.\ MPPI rollouts.}
    \label{fig:maze2d_qual}
    \label{fig:maze2d_SR_vs_numsamples}
    \vspace{-1.6em}
\end{figure*}

\textbf{Results.}
As shown in Fig.~\ref{fig:maze2d_SR_vs_numsamples} (right), our method achieves the highest SR and shortest average path length, outperforming all baselines across all rollout budgets. Vanilla MPPI and Log-MPPI reach reasonable success rates but often waste rollouts in concave regions, trapping the planner in local minima (Fig.~\ref{fig:maze2d_qual}, left-a).
Diff-MPPI partially addresses this by using the diffusion trajectory as a global nominal and applying MPPI for local refinement, but its fixed proposal covariance creates a single-scale trade-off: too small and MPPI cannot escape local basins or repair errors in the learned prior; too large and many rollouts collide near walls, leaving only a few near-boundary survivors to dominate the MPPI update. The resulting MPPI trajectory tends to skim obstacles and becomes more collision-prone.
Our method overcomes this limitation by leveraging the tube uncertainty from \mymethod{} as a flexibility-aware variance: the proposal expands in open regions to encourage exploration and contracts in tight passages to suppress wall-penetrating samples (Fig.~\ref{fig:maze2d_qual}, left-b).
This lets MPPI efficiently refine the diffusion prior into a collision-free trajectory.

\subsection{Case Study 2: Y-Corridor Pedestrian Navigation}\label{exp:y_corridor}
Next, to evaluate whether the uncertainty produced by \mymethod{} translates into safer downstream control (Q3), we deploy our method as a ``world model'' (more precisely, a pedestrian motion predictor) within an MPPI controller on a Y-shaped junction corridor benchmark (in Sec. \ref{exp:ph_property}).
In this ORCA-simulated environment, a robot navigates to a goal while avoiding $N_p\in\{3,6,9,12\}$ pedestrians with randomized starts and goals.
Pedestrians split between the left and right corridors, creating multi-modal futures and dense congestion at the junction center.
MPPI evaluates candidate rollouts at each control step, with the pedestrian forecast and its uncertainty fed into a chance-cost formulation (see Appendix~\ref{sec:gmm_chance}).
We compare against three baselines:
\textbf{NF} (No Future), which runs MPPI without any prediction;
\textbf{LKF}, which feeds a Linear Kalman Filter forecast into MPPI~\citep{mohamed2025c2umppi};
and \textbf{MID-Ensemble}, which draws multiple diffusion samples per pedestrian and fits a Gaussian mixture to the resulting set for chance-cost evaluation.
We further ablate three design choices: \textbf{MID-UM} uses a single unimodal diffusion prediction.
\textbf{MID-MM} uses multimodal predictions but without \mymethod{},
and \textbf{Ours-noCal} uses the full pipeline but skips post-hoc calibration.

\textbf{Results.}
Table~\ref{tab:y_corridor} shows that \mymethod{} achieves the best or tied-best SR across all crowd densities (with notably better results at $N_p = 9$ and $N_p = 12$).
Without any prediction (NF), the planner is purely reactive. Adding even a constant-velocity forecast (LKF) substantially raises SR, confirming that anticipating pedestrian motion is critical.
Diffusion-based predictions (MID-UM/MM) further help but remain brittle: unimodal commits to a single future, while multimodal lacks calibrated per-timestep uncertainty to modulate conservatism.
MID-Ensemble estimates uncertainty by brute-force sampling, yet its heavy computation limits the control frequency, and SR remains below that of our method across all densities.
\mymethod{} yields the best SR without sacrificing efficiency. Navigation time $\overline{T}$ and path length $\overline{d}$ remain comparable to the strong baselines.
Comparing Ours-noCal to Ours, skipping calibration consistently lowers SR in dense crowds, confirming that calibrated uncertainty is important for chance-cost control.

\begin{table}[t]
    \centering
    \caption{Y-corridor navigation results across crowd densities $N_p\in\{3,6,9,12\}$.}
    \label{tab:y_corridor}
    \small
    \setlength{\tabcolsep}{3pt}
    \resizebox{\textwidth}{!}{
    \begin{threeparttable}

  \begin{tabular}{l|cccc|cccc|cccc}
            \toprule                                                                                                                 
            & \multicolumn{4}{c|}{\textbf{SR [\%] $\uparrow$}} & \multicolumn{4}{c|}{\textbf{$\overline T$ [s]}} &                   
  \multicolumn{4}{c}{\textbf{$\overline d$                                                                                           
     [m]}} \\                                                                                                                        
            \textbf{Method} & 3 & 6 & 9 & 12 & 3 & 6 & 9 & 12 & 3 & 6 & 9 & 12 \\                                                    
            \midrule                                                                                                                 
            NF & 80.0 & 72.0 & 70.0 & 65.0 & \textbf{14.44{\scriptsize$\pm$3.95}} & \textbf{10.05{\scriptsize$\pm$6.91}} &           
  \textbf{8.91{\scriptsize$\pm$7.53}}                                                                                                
     & \textbf{8.82{\scriptsize$\pm$7.63}} & \textbf{10.14{\scriptsize$\pm$2.75}} & \textbf{6.88{\scriptsize$\pm$4.87}} &
    \textbf{5.97{\scriptsize$\pm$5.20}} & \textbf{6.02{\scriptsize$\pm$5.37}} \\                                                     
            LKF & 96.0 & 91.0 & \underline{90.0} & 85.0 & 14.64{\scriptsize$\pm$3.65} & 11.49{\scriptsize$\pm$6.92} &
    10.28{\scriptsize$\pm$7.50} & 10.68{\scriptsize$\pm$8.87} & 10.26{\scriptsize$\pm$2.51} & 7.95{\scriptsize$\pm$4.92} &           
  6.99{\scriptsize$\pm$5.21} &                                                                                                       
    7.29{\scriptsize$\pm$6.06} \\                                                                                                    
            MID-UM\tnote{*} & \textbf{100.0} & 89.0 & 85.0 & 86.0 & 14.72{\scriptsize$\pm$3.98} & 11.95{\scriptsize$\pm$7.68} &      
  11.19{\scriptsize$\pm$9.59}                                                                                                        
     & 11.84{\scriptsize$\pm$10.13} & 10.31{\scriptsize$\pm$2.77} & 8.24{\scriptsize$\pm$5.40} & 7.64{\scriptsize$\pm$6.64} &        
  8.09{\scriptsize$\pm$6.83} \\                                                                                                      
            MID-MM\tnote{*} & \textbf{100.0} & 91.0 & 85.0 & 88.0 & 14.74{\scriptsize$\pm$3.98} & 12.08{\scriptsize$\pm$7.47} &
    10.56{\scriptsize$\pm$8.18} & 11.25{\scriptsize$\pm$9.84} & 10.30{\scriptsize$\pm$2.75} & 8.29{\scriptsize$\pm$5.22} &           
  7.21{\scriptsize$\pm$5.76} &                                                                                                       
    7.71{\scriptsize$\pm$6.68} \\                                                                                                    
            MID-Ens-64 & 98.0 & 87.0 & 82.0 & \underline{89.0} & 14.55{\scriptsize$\pm$3.91} & \underline{11.43{\scriptsize$\pm$7.22}} &         
    9.89{\scriptsize$\pm$7.72} & \underline{10.59{\scriptsize$\pm$8.64}} & 10.19{\scriptsize$\pm$2.71} &                             
  \underline{7.89{\scriptsize$\pm$5.13}} &                                                                                           
    6.73{\scriptsize$\pm$5.41} & \underline{7.24{\scriptsize$\pm$5.97}} \\                                                           
            MID-Ens-128 & \underline{99.0} & 90.0 & 82.0 & {86.0} & 14.60{\scriptsize$\pm$3.97} &                          
  11.68{\scriptsize$\pm$7.25} &                                                                                                      
    \underline{9.57{\scriptsize$\pm$7.42}} & 10.95{\scriptsize$\pm$8.66} & 10.21{\scriptsize$\pm$2.76} & 8.06{\scriptsize$\pm$5.14} &
    \underline{6.52{\scriptsize$\pm$5.19}} & 7.50{\scriptsize$\pm$5.95} \\                                                           
            MID-Ens-256 & \underline{99.0} & 90.0 & 83.0 & 87.0 & 14.57{\scriptsize$\pm$3.91} & 11.53{\scriptsize$\pm$6.96} &        
    10.04{\scriptsize$\pm$7.85} & 10.71{\scriptsize$\pm$8.91} & 10.19{\scriptsize$\pm$2.74} & 7.95{\scriptsize$\pm$4.94} &           
    6.86{\scriptsize$\pm$5.47} & 7.30{\scriptsize$\pm$6.07} \\                                                                       
            Ours-noCal & \textbf{100.0} & \underline{94.0} & 86.0 & 88.0 & \underline{14.50{\scriptsize$\pm$3.86}} &                 
  11.89{\scriptsize$\pm$7.34} &                                                                                                      
    10.27{\scriptsize$\pm$7.75} & 11.35{\scriptsize$\pm$9.34} & \underline{10.18{\scriptsize$\pm$2.69}} & 8.21{\scriptsize$\pm$5.18} 
  & 7.02{\scriptsize$\pm$5.45} &                                                                                                     
    7.75{\scriptsize$\pm$6.40} \\
            Ours & \textbf{100.0} & \textbf{96.0} & \textbf{93.0} & \textbf{92.0} & 14.65{\scriptsize$\pm$3.97} &                    
  11.96{\scriptsize$\pm$7.08} &                                                                                                      
    11.71{\scriptsize$\pm$9.34} & 11.99{\scriptsize$\pm$10.25} & 10.30{\scriptsize$\pm$2.76} & 8.32{\scriptsize$\pm$5.04} &          
  7.99{\scriptsize$\pm$6.49} &                                                                                                       
    8.19{\scriptsize$\pm$6.96} \\
            \bottomrule                                                                                                              
        \end{tabular}

    \begin{tablenotes}[flushleft]\footnotesize
        \item[*] Without uncertainty tube (\mymethod{}). SR: success rate, $\overline T$: avg.\ navigation time (successful episodes only), $\overline d$: avg.\ path length (successful episodes only). \textbf{Bold}: best, \underline{underline}: second best.
    \end{tablenotes}
    \end{threeparttable}}
\vspace{-1.8em}
\end{table}

\subsection{Case Study 3: Real-Robot Manipulation}\label{Sec: Real-robot}
Finally, we evaluate \mymethod{} on a 7-DOF Franka Emika arm in a tabletop pick-and-place task (Fig.~\ref{fig:teaser}, bottom). The arm must reach inside a shelf to pick the tomato and place it on the upper shelf, creating a constrained non-convex workspace. The box obstacles are randomly repositioned across trials. The diffusion model provides two approach modes with different clearance profiles, a flexible front approach and a tighter side approach. We test two conditions. \textbf{Static} evaluates policy-side tube-informed MPPI with 500 rollouts and stationary obstacles. \textbf{Dynamic} adds a human reaching for a nearby object and evaluates whether \mymethod{}-based motion prediction improves avoidance. Each method uses 30 trials per condition. We report success rate (SR), collision rate (CR), and timeout rate (TO), where TO denotes collision-free failure within 120\,s. More details in Appendix~\ref{app:exp_real_robot}.

\begin{wraptable}{r}{0.26\textwidth}
    \vspace{-13pt}
    \raggedleft
    \scriptsize
    \caption{Static setting.}
    \vspace{-6pt}
    \label{tab:real_robot_static}
    \setlength{\tabcolsep}{3pt}
    \begin{tabular}{@{}lccc@{}}
        \toprule
        \textbf{Method} & SR$\uparrow$ & CR$\downarrow$ & TO$\downarrow$ \\
        \midrule
        Diffuser     & 0.47 & 0.53 & 0.0 \\
        Vanilla MPPI & 0.17 & 0.80 & 0.03 \\
        Diff-MPPI    & 0.50 & 0.23 & 0.27 \\
        Ours         & \textbf{0.83} & \textbf{0.13} & \textbf{0.03} \\
        \bottomrule
    \end{tabular}
    \vspace{-10pt}
\end{wraptable}

\textbf{Static setting} (Table~\ref{tab:real_robot_static}).
Diffuser executes open-loop trajectories and frequently collides with the obstacle.
After the pick, Vanilla MPPI often gets trapped between shelf levels because its short horizon cannot sample a motion that exits the shelf before moving to the upper target level, leading to upper-shelf collision.
Diff-MPPI’s diffusion prior guides the arm toward the goal, but without motion-flexibility awareness, it may choose the tighter side approach. Fixed per-joint exploration can then drive the arm into an unfavorable posture near the shelf, where joint limits and obstacle constraints create a local minimum. Short-horizon MPPI may remain stuck there until timeout (TO), or collide with the shelf while attempting to recover.
\mymethod{} uses the predicted tube to select more flexible modes and adapt exploration along the selected path, yielding the highest SR and the lowest CR.

\begin{wraptable}{r}{0.36\textwidth}
    \vspace{-13pt}
    \raggedleft
    \scriptsize
    \caption{Dynamic setting.}
    \label{tab:real_robot_dynamic}
    \vspace{-4.5pt}
    \setlength{\tabcolsep}{3pt}
    \begin{tabular}{@{}lcccc@{}}
        \toprule
        \textbf{Method} & SR$\uparrow$ & CR\textsubscript{env}$\downarrow$ & CR\textsubscript{human}$\downarrow$ & TO$\downarrow$ \\
        \midrule
        Ours (no WM) & 0.37 & 0.27 & 0.40 & \textbf{0.0} \\
        MID-Ens-64  & 0.53 & 0.27 & 0.27 & \textbf{0.00} \\
        MID-Ens-128 & 0.47 & 0.30 & 0.33 & \textbf{0.00} \\
        Ours         & \textbf{0.70} & \textbf{0.13} & \textbf{0.10} & 0.07 \\
        \bottomrule
    \end{tabular}

    \vspace{2pt}
    {\tiny A single episode may incur both collision types.}
    \vspace{-10pt}
\end{wraptable}
\textbf{Dynamic setting} (Table~\ref{tab:real_robot_dynamic}).
Our full method adds \mymethod{}-based human motion prediction to the same tube-informed MPPI controller. Without prediction, the planner reacts only when the human is already close, which leads to late evasion or over-correction into the shelf. Sampling-based prediction faces a trade-off between uncertainty quality and response speed. MID-Ens-64 is the largest ensemble that can provide a fresh prediction every control cycle, yet remains less successful than \mymethod{}. Increasing the sample count to MID-Ens-128 slows prediction updates and delays the controller's response to human motion. With \mymethod{}, the arm anticipates the human's future occupancy from the predicted tube and yields earlier, achieving higher success and fewer human collisions than both ensemble baselines. Remaining failures of \mymethod{} are due to imperfect human occupancy prediction or yielding into low-flexibility configurations near the shelf, where the short-horizon controller cannot recover. Qualitative case studies are provided in Appendix~\ref{app:real_robot_cases}.

\section{Conclusion} 
We presented \mymethod{}, a score-curvature-based uncertainty module that turns diffusion trajectory models into control-ready Gaussian tubes and improves closed-loop planning across forecasting, navigation, and real-robot manipulation. SCOPE  makes diffusion uncertainty directly actionable for real-time robot control.

\section{Limitations} 
\vspace{-.5em}
\mymethod{} assumes a reliable diffusion backbone, since its uncertainty estimate is distilled from the learned score. When training data has limited coverage or the diffusion model fails to learn separate modes, the predicted tubes may be miscalibrated. Our current experiments use relatively low-dimensional state and geometry inputs. Extending \mymethod{} to raw visual observations, such as images or point clouds, is an important direction for future work.

\acknowledgments{
This research / project is supported by the Ministry of Education, Singapore, under its Academic Research Fund Tier 2 (T2EP20225-0029) and Academic Research Fund Tier 1 (T1 251RES2515). 
This work is also partly supported by Défi Clé Robotique centrée sur l'Humain funded by Région Occitanie and by the AS2 of PEPR O2R (ANR-22-EXOD-0006), France.
}

\bibliography{Bib}  %
\clearpage
\appendix
\section{Related Works}\label{Sec: Related Works}

\subsection{Diffusion models for trajectory prediction.}
Diffusion models have become a prominent generative framework for motion forecasting and planning due to their ability to represent complex multimodal distributions. As \emph{world models}, they forecast the future trajectories of surrounding agents: 
MID~\citep{gu2022stochastictrajectorypredictionmotion} introduces diffusion to capture motion indeterminacy and subsequent work extends diffusion forecasting to multi-agent settings~\citep{jiang2023motiondiffuser}, faster sampling~\citep{mao2023leapfrogdiffusionmodelstochastic}, and intention conditioning~\citep{liu2025intention}. As policies, diffusion models generate robot trajectories or action sequences directly: Diffuser~\citep{janner2022planning} and Decision Diffuser~\citep{ajay2023conditionalgenerativemodelingneed} formulate planning as conditional trajectory denoising, Diffusion Policy~\citep{chi2024diffusionpolicyvisuomotorpolicy} learns visuomotor policies by denoising action sequences conditioned on observations, while G-SSIP~\citep{jiang2026guided} introduces guided streaming generation, allowing the policy to remain reactive to environmental changes during execution, and PB-Diff~\citep{luo2024potentialbaseddiffusionmotion} learns composable potential functions for constrained motion planning.
Despite strong coverage of multimodal futures, most of these models do not expose a structured joint trajectory precision, or fast per-timestep marginal covariance interface, that downstream chance-constrained controllers can consume directly.

\subsection{Uncertainty quantification from generative trajectory models.}
Uncertainty in trajectory models is commonly separated into epistemic and aleatoric components. Epistemic uncertainty reflects uncertainty about the learned model or its parameters and can indicate limited data coverage, model misspecification, or out-of-distribution inputs~\cite{valdenegrotoro2022deeperlookaleatoricepistemic}. Ensemble and Bayesian-style methods estimate such model confidence in forecasting and generative models~\cite{eltouny_-tgn_2023, jazbec2025generativeuncertaintydiffusionmodels,shu_zero-shot_2024}.

For online control, the immediate object needed by MPPI and chance-constrained planning is often aleatoric, or predictive, uncertainty over future trajectories around each plausible mode. Diffusion models represent this uncertainty implicitly through generated samples. A standard practical approach is therefore to draw many futures and compute dispersion statistics or fit a Gaussian/GMM over the resulting trajectories~\cite{neumeier2026uncertaintyawarediffusionmodelmultimodal,wintel2025usingdiffusionensemblesestimate}. Related trajectory-diffusion methods also learn state-wise uncertainty or uncertainty-aware refinements~\cite{capellera2025unifieduncertaintyawarediffusionmultiagent}. U2Diff provides state-wise uncertainty and mode-level error ranking for trajectory completion, but its uncertainty is per-state and ranking-oriented rather than a structured joint trajectory precision designed for closed-loop control. Sampling-based and ensemble-based estimates, on the other hand, require repeated diffusion decoding or many generated trajectories, which can be too slow for closed-loop control. Learned reverse-process variances in DDPMs~\cite{nichol2021improveddenoisingdiffusionprobabilistic} are also related, but they describe denoising transition uncertainty rather than terminal trajectory covariance or log-density curvature around a decoded clean trajectory. Conformal methods provide calibrated coverage sets or tubes for planning and prediction~\cite{sun2023conformal, muthali2023multiagentreachabilitycalibrationconformal}. While these sets offer valuable coverage guarantees, they do not directly expose a structured Gaussian precision or covariance representation with fast marginal queries for chance-constrained control. SCOPE targets this gap by producing a compact, calibrated, per-mode structured trajectory precision whose marginal covariances can be queried efficiently for control.

\subsection{Structured uncertainty for sampling-based control.}
Sampling-based controllers such as MPPI optimize controls through many stochastic rollouts and importance-weighted updates~\cite{dramppi-21,williams2017mppi, alvarez-padilla_real-time_2024}, naturally handling nonconvex costs without linearization~\citep{c2umppi-5,c2umppi-6,c2umppi-7}. Recent robotic systems couple online perception with reactive planning in dynamic environments~\citep{hu2026visibilityawaremobilegraspingdynamic}. In human-robot settings, COSMIK-MPPI~\cite{gursoy2026cosmikmppiscalingconstrainedmodel} combines markerless human motion estimation with constraint-aware rollout termination, showing that MPPI can reactively handle close-proximity humans, but it does not rely on explicit human motion prediction. This makes prediction uncertainty an important missing interface when the robot must anticipate future human occupancy rather than only react to the current estimate.

Chance-constrained and risk-aware MPPI methods make this predictive interface explicit. C2U-MPPI~\citep{mohamed2025c2umppi} integrates deterministic reformulations of probabilistic constraints into unscented MPPI~\citep{c2umppi-12} for real-time obstacle avoidance~\citep{c2umppi-10,c2umppi-11,c2umppi-15} and incorporates probabilistic collision constraints using lightweight Linear-Kalman-filter predictions, which are efficient but limited for multimodal future motion. DRA-MPPI~\cite{trevisan2025dynamicriskawaremppimobile} efficiently evaluates joint collision probabilities under stochastic obstacle predictions such as Gaussian mixtures, but it assumes these predictive distributions are provided by an external or task-specific motion model. These works therefore leave open a complementary problem: producing an expressive diffusion-level predictive uncertainty in a compact form suitable for real-time MPPI. SCOPE addresses this by augmenting each diffusion mode into a calibrated, structured precision tube that supports fast marginal covariance queries across rollouts, obstacles, and timesteps.

Structured Gaussian and precision representations provide a natural way to realize this interface. In trajectory problems, sparse precision matrices encode temporal conditional independence and enable efficient inference, a principle exploited by Gaussian-process motion planning and its differentiable extensions~\cite{Mukadam_2018,bhardwaj_differentiable_2020}. Learned structured uncertainty heads further show that neural networks can predict positive-definite uncertainty operators beyond diagonal covariance~\cite{Dorta_2018_CVPR}. SCOPE builds on this structured-precision view, but differs in how the uncertainty is obtained: it distills terminal score curvature from a diffusion trajectory model into a structured precision head using HVP/Hutchinson supervision~\cite{pearlmutter1994fast,hutchinson1989stochastic}. The resulting calibrated Gaussian tubes preserve trajectory structure while exposing the marginal covariances needed by MPPI and chance-constrained controllers.

\section{Method Details}\label{Section: Method Details}

\subsection{Laplace's Method}\label{app:laplace_approximation}

Let $\y \in \mathbb{R}^{dT}$ denote the future trajectory, let $\mathbf{c}$ denote the conditioning context, and let $z \in [K]$ denote a discrete latent mode. We write the predictive distribution as a mixture over modes:
\begin{align}
    p(\y \mid \mathbf{c})
    &= \sum_{k=1}^{K} p(\y, z{=}k \mid \mathbf{c}) \\
    &= \sum_{k=1}^{K} p(z{=}k \mid \mathbf{c})\, p(\y \mid \mathbf{c}, z{=}k) \\
    &\approx \sum_{k=1}^{K} \pi_k \cdot \mathcal{N}(\boldsymbol\mu_k, \boldSigma_k).
\end{align}
The final equality relies on the approximation that each mode-conditional density $p(\y \mid \mathbf{c}, z{=}k)$ is locally Gaussian. We justify this approximation using Laplace's method.

\paragraph{Local Taylor Expansion of the Log-Density.}

Fix a mode $k$ and define the mode-conditional log-density
\[
    \ell(\y) := \log p(\y \mid \mathbf{c}, z{=}k).
\]
Assume that $\ell$ is twice continuously differentiable. Let
\[
    \boldsymbol\mu_k := \arg\max_{\y} \ell(\y)
\]
be the $k$-th mode. Since $\boldsymbol\mu_k$ is a local maximizer, the first-order optimality condition gives
\begin{equation}
    \nabla_\y \ell(\y)\big|_{\y = \boldsymbol\mu_k} = \mathbf{0}.
\end{equation}

Now expand $\ell$ to second order around $\boldsymbol\mu_k$. Writing
\[
    \boldsymbol\delta = \y - \boldsymbol\mu_k,
\]
we obtain
\begin{align}
    \ell(\boldsymbol\mu_k + \boldsymbol\delta)
    &=
    \ell(\boldsymbol\mu_k)
    +
    (\underbrace{\nabla_\y \ell(\boldsymbol\mu_k)}_{= \mathbf{0}})^\top \boldsymbol\delta
    +
    \frac{1}{2}
    \boldsymbol\delta^\top
    \nabla_\y^2 \ell(\boldsymbol\mu_k)
    \boldsymbol\delta
    +
    {o}(\|\boldsymbol\delta\|^2)\\
    &\approx
    \ell(\boldsymbol\mu_k)
    +
    \frac{1}{2}
    \boldsymbol\delta^\top
    \nabla_\y^2 \ell(\boldsymbol\mu_k)
    \boldsymbol\delta.
\end{align}

\paragraph{Recovering the Gaussian Approximation.}

We define the local precision matrix as the negative Hessian of the log-density evaluated at the mode:
\begin{equation}
    \boldsymbol\Lambda^*
    :=
    -\nabla_\y^2 \ell(\boldsymbol\mu_k)
    =
    -\nabla_\y^2
    \log p(\y \mid \mathbf{c}, z{=}k)
    \big|_{\y = \boldsymbol\mu_k}.
\end{equation}
At a local maximum, the Hessian is negative semi-definite, and therefore $\boldsymbol\Lambda^*$ is positive semi-definite. We assume that the mode is non-degenerate, so that $\boldsymbol\Lambda^*$ is strictly positive definite.

Substituting
\[
    \nabla_\y^2 \ell(\boldsymbol\mu_k) = -\boldsymbol\Lambda^*
\]
into the quadratic approximation yields
\begin{equation}
    \log p(\y \mid \mathbf{c}, z{=}k)
    \approx
    \ell(\boldsymbol\mu_k)
    -
    \frac{1}{2}
    (\y - \boldsymbol\mu_k)^\top
    \boldsymbol\Lambda^*
    (\y - \boldsymbol\mu_k).
\end{equation}
Exponentiating both sides gives
\begin{equation}
    p(\y \mid \mathbf{c}, z{=}k)
    \approx
    \exp\!\big(\ell(\boldsymbol\mu_k)\big)
    \exp\!\left(
    -\frac{1}{2}
    (\y - \boldsymbol\mu_k)^\top
    \boldsymbol\Lambda^*
    (\y - \boldsymbol\mu_k)
    \right).
\end{equation}
Thus, up to a normalising constant, the mode-conditional density has the form of a Gaussian distribution with mean $\boldsymbol\mu_k$ and precision matrix $\boldsymbol\Lambda^*$. Equivalently, its covariance is
\[
    \boldSigma_k = (\boldsymbol\Lambda^*)^{-1}.
\]
Therefore,
\begin{equation}
    p(\y \mid \mathbf{c}, z{=}k)
    \approx
    \mathcal{N}\!\big(\y;\, \boldsymbol\mu_k,\, (\boldsymbol\Lambda^*)^{-1}\big).
\end{equation}

\paragraph{Anchor and curvature diagnostics.}\label{app:anchor_diagnostics}
To test the local Gaussian approximation, we evaluate the curvature of the diffusion score at predicted anchors.
We evaluate 1,000 focal-pedestrian predictions from 20 held-out episodes in the 3-pedestrian Y-corridor setting, retaining the top three of six modes to obtain 3,000 anchors.
At each anchor $a$, let $g$ denote the local score and $C$ the symmetrized negative score Jacobian.
Of the 3,000 raw curvature matrices, 2,843 (94.8\%) are positive definite.
For these anchors, the local quadratic center is $a+C^{-1}g$.
The mode-probability-weighted median RMS positional shift to this center is 1.33~cm.

We separately assess covariance sensitivity using the learned precision $\Lambda_\phi$, which defines a quadratic center $a+\Lambda_\phi^{-1}g$.
Holding mixture means fixed at this center, we query covariance at the center and the decoded anchor.
Moving the query to the anchor increases mixture NLL by only 0.027.
This indicates that uncertainty quality changes little at the observed anchor offset.

\subsection{Markov Sparsity in Trajectory Precision}\label{app:markov_sparsity}

We motivate the use of a block-banded sparsity pattern for the trajectory precision matrix.

\vspace{1em}

Consider a trajectory $\y_{1:T}$ generated by a dynamical process in which each future state depends only on a limited temporal neighbourhood, together with process noise. Under a $b$-th order Markov assumption, the joint density can be factorised as
\begin{equation}
    p(\y_{1:T})
    =
    p(\y_1)
    \prod_{t=2}^{T}
    p\!\left(
        \y_t
        \mid
        \y_{t-1}, \ldots, \y_{\max(1,t-b)}
    \right),
\end{equation}
i.e., each timestep depends directly only on the previous $b$ timesteps.

\vspace{1em}

For Gaussian graphical models, the sparsity pattern of the precision matrix encodes conditional independence relationships. In particular, if
\[
    \y \sim \mathcal{N}(\boldmu, \boldSigma),
\]
with precision matrix
\[
    \boldsymbol\Lambda = \boldsymbol\Sigma^{-1},
\]
then
\begin{equation}
    \boldsymbol\Lambda_{ij} = 0
    \quad \Longleftrightarrow \quad
    \mathbf y_i \perp \mathbf y_j
    \mid
    \mathbf y_{\setminus\{i,j\}}.
\end{equation}
Thus, zero entries in the precision matrix correspond to pairs of variables that are conditionally independent given the remaining variables.

\vspace{1em}

For a $b$-th order Markov trajectory model, timesteps $i$ and $j$ do not directly interact once the intervening states are conditioned on, whenever
\[
    |i-j| > b.
\]
Consequently, the corresponding precision blocks satisfy
\begin{equation}
    \boldsymbol\Lambda_{ij} = 0
    \quad \text{for} \quad
    |i-j| > b.
\end{equation}
This yields a block-banded precision matrix with temporal bandwidth $b$.

We further use a block-banded precision to model local temporal dependencies. To accommodate dependencies beyond this local Markov structure, we augment the banded precision with a learned low-rank term $\mathbf R\mathbf R^\top$. This term allows additional couplings between distant timesteps while retaining efficient inference.

\begin{figure}[t]
    \centering
    \includegraphics[width=\linewidth]{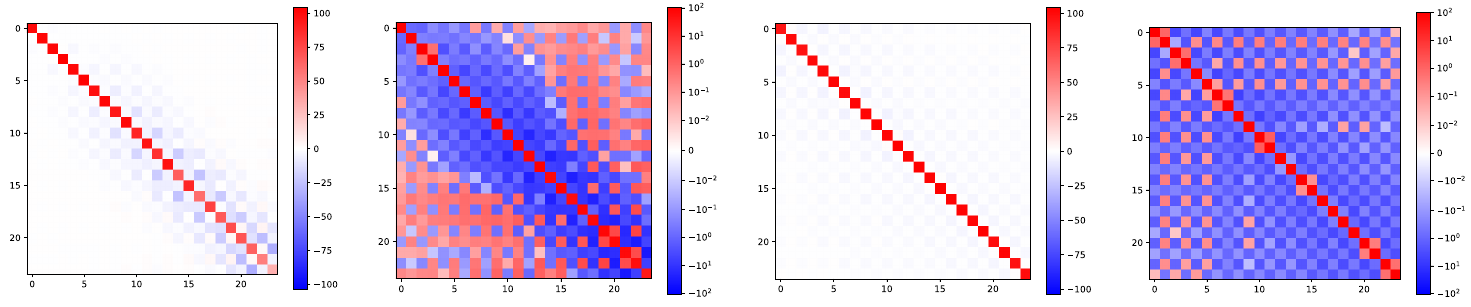}
\caption{
Empirical structure of teacher precision matrices.
The left two plots and the right two plots correspond to two representative trajectories, respectively.
For each trajectory, we visualize the same precision matrix with a linear scale and a symmetric log scale.
The linear-scale views show strong diagonal dominance, while the log-scale views reveal banded local structure and off-band global patterns.
}
    \label{fig:teacher_precision_examples}
\end{figure}
\paragraph{Empirical structure of teacher precision.}
Fig.~\ref{fig:teacher_precision_examples} visualizes teacher precision matrices
computed from diffusion score curvature for two representative trajectories.
For each trajectory, we show the same matrix under a linear color scale and a
symmetric logarithmic color scale. The linear-scale views show strong diagonal
dominance, while the log-scale views reveal a clear near-diagonal banded
structure together with structured off-band global patterns. This empirical
structure supports our parameterization: the block-banded SPD term captures the
dominant local temporal coupling, and the rank-$r$ low-rank correction captures
residual global correlations induced by goals, obstacles, and scene context.

\subsection{Banded Cholesky Structure}\label{app:chol_structure}

We summarise the structural relationship between block-banded symmetric positive definite (SPD) matrices and their Cholesky factors. This relationship is what makes our precision parameterisation computationally efficient.

Let the full trajectory dimension be $D = Td$, where $T$ is the prediction horizon and $d$ is the state dimension at each timestep. We view a matrix $\boldsymbol\Lambda \in \mathbb{R}^{D \times D}$ as a $T \times T$ block matrix with blocks of size $d \times d$.

\textbf{Fact.} Let $\boldsymbol\Lambda$ be SPD and block-banded with block bandwidth $b$, meaning that
\[
    \boldsymbol\Lambda_{ij} = 0
    \quad \text{whenever} \quad
    |i-j| > b,
\]
where $\boldsymbol\Lambda_{ij} \in \mathbb{R}^{d \times d}$ denotes the $(i,j)$-th block. Then its Cholesky factor $\mathbf L$, defined by
\[
    \boldsymbol\Lambda = \mathbf L\mathbf L^\top,
\]
is also block lower-banded with bandwidth $b$:
\[
    \mathbf L_{ij} = 0
    \quad \text{whenever} \quad
    i-j > b.
\]
Therefore, by parameterising
\[
    \mathbf A_\phi = \mathbf L_\phi \mathbf L_\phi^\top
\]
with a block lower-banded Cholesky factor $\mathbf L_\phi$, we ensure that $\mathbf A_\phi$ is an SPD block-banded matrix. Conversely, any block-banded SPD matrix admits a Cholesky factor with the same block bandwidth. Hence, within the family of block-banded SPD matrices, this parameterisation is not restrictive.

\textbf{Computational benefit.}
Exploiting the banded structure substantially reduces the cost of factorisation. For a block matrix with $T$ temporal blocks, block size $d \times d$, and block bandwidth $b$, banded Cholesky factorisation costs
\[
    O(T b^2 d^3),
\]
whereas dense Cholesky factorisation of the full $D \times D$ matrix costs
\[
    O(D^3) = O(T^3 d^3).
\]
Thus, for fixed $d$ and $b$, the factorisation cost scales linearly rather than cubically with the prediction horizon $T$. The same sparsity structure also enables efficient triangular solves and selected inverse computations, such as the extraction of marginal covariance blocks.

\textbf{Derivation of factorisation cost.}
At each temporal block column $t$ of the block Cholesky factorisation, the Schur complement update involves interactions among at most $b$ previously computed neighbouring blocks and at most $b$ target sub-diagonal blocks. This gives $O(b^2)$ block operations per temporal column. Each block operation is a $d \times d$ matrix product with cost $O(d^3)$. Over $T$ temporal columns, the total factorisation cost is therefore
\[
    O(T b^2 d^3).
\]

\subsection{Inference Complexity Analysis}\label{app:inference_complexity}

At inference time, the precision head predicts the parameters of the structured precision matrix
\[
    \boldsymbol\Lambda_\phi = \mathbf A_\phi + \mathbf R \mathbf R^\top,
\]
where $\mathbf A_\phi = \mathbf L_\phi \mathbf L_\phi^\top$ is block-banded SPD and $\mathbf R \in \mathbb{R}^{Td \times r}$ is a low-rank factor with rank $r$. Our goal is to recover the per-timestep marginal covariance blocks
\[
    \boldSigma_{tt} \in \mathbb{R}^{d \times d}
\]
from the covariance matrix
\[
    \boldSigma = \boldsymbol\Lambda_\phi^{-1}.
\]

\textbf{Step 1: Reuse the predicted Cholesky factor.}
The precision head directly predicts the block lower-banded factor $\mathbf L := \mathbf L_\phi$, with
\[
    \mathbf A_\phi = \mathbf L \mathbf L^\top.
\]
We reuse this factor for the structured covariance queries.

\textbf{Step 2: Low-rank Woodbury correction.}
The full precision matrix includes the low-rank correction
\[
    \boldsymbol\Lambda_\phi = \mathbf A_\phi + \mathbf R \mathbf R^\top.
\]
Applying the Woodbury identity gives
\begin{equation}
    \boldsymbol\Lambda_\phi^{-1}
    =
    \mathbf A_\phi^{-1}
    -
    \mathbf A_\phi^{-1}\mathbf R
    \left(
        \mathbf I_r + \mathbf R^\top \mathbf A_\phi^{-1}\mathbf R
    \right)^{-1}
    \mathbf R^\top \mathbf A_\phi^{-1}.
\end{equation}
The term $\mathbf A_\phi^{-1}\mathbf R$ is computed by solving against $\mathbf R$ using the banded triangular factors $\mathbf L$ and $\mathbf L^\top$. Since $\mathbf R$ has $r$ right-hand sides, and each banded triangular solve processes $T$ block rows with at most $b$ neighbouring block interactions per row, this costs
\[
    O(T b d^2 r).
\]
The Woodbury core matrix
\[
    \mathbf M
    =
    \mathbf I_r + \mathbf R^\top \mathbf A_\phi^{-1}\mathbf R
\]
has size $r \times r$, so forming and inverting it costs
\[
    O(T d r^2 + r^3),
\]
where the first term accounts for forming $\mathbf R^\top \mathbf A_\phi^{-1}\mathbf R$ and the second term accounts for the dense $r \times r$ inverse.

\textbf{Step 3: Extracting marginal covariance blocks.}
To obtain the diagonal blocks of
\[
    \boldSigma = \boldsymbol\Lambda_\phi^{-1},
\]
we combine the selected diagonal blocks of $\mathbf A_\phi^{-1}$ with the Woodbury correction. The required diagonal blocks of $\mathbf A_\phi^{-1}$ can be computed efficiently using selected inversion methods, such as Takahashi's equations, at cost
\[
    O(T b^2 d^3).
\]
Let
\[
    \mathbf B = \mathbf A_\phi^{-1}\mathbf R.
\]
For timestep $t$, let $\mathbf B_t \in \mathbb{R}^{d \times r}$ denote the block rows of $\mathbf B$ corresponding to that timestep. The Woodbury correction to the marginal covariance block is
\[
    \mathbf B_t \mathbf M^{-1} \mathbf B_t^\top.
\]
Computing this correction for all $T$ marginal blocks costs
\[
    O(T d r^2 + T d^2 r).
\]

\textbf{Overall complexity.}
For $b\geq 1$, combining the triangular solves, Woodbury core computation, selected inversion, and marginal block extraction gives a total cost of
\[
    O\!\left(
        T b^2 d^3
        + T b d^2 r
        + T d r^2
        + r^3
    \right).
\]
Thus, for fixed block dimension $d$, bandwidth $b$, and low-rank dimension $r$, marginal covariance extraction using this structured algorithm scales linearly with the prediction horizon $T$.

\textbf{Implementation note.}
The algorithmic interface is compatible with block-banded Cholesky factorisation, selected inversion, and Woodbury updates. In our reported experiments, however, the trajectory dimension is small: $d{=}2$, $T{=}12$, and hence $D = Td = 24$. Our implementation uses dense Cholesky factorisation and triangular solves, which complete in under $0.5\,\mathrm{ms}$ on GPU. The banded complexity above describes the scalable implementation path for substantially longer prediction horizons.

\subsection{Mixture-of-Diffusion Training Details}\label{app:mixture_training}
We derive the ELBO for the discrete-latent conditional diffusion model (Sec.~\ref{sec:method_mixture}). We write $\mathbf y_0\in\mathbb{R}^{dT}$ for the clean trajectory and $\mathbf y_{1:\mathsf S}$ for the diffusion-noised latents.

\paragraph{Generative model.}
The joint distribution over trajectory $\mathbf y_0$, diffusion latents $\mathbf y_{1:\mathsf S}$, and mode $z$ given context $\mathbf c$ is:
\begin{equation}
p(\mathbf y_0, \mathbf y_{1:\mathsf S}, z \mid \mathbf c) = p_\theta(z\mid\mathbf c)\, p(\mathbf y_\mathsf S)\prod_{s=1}^{\mathsf S} p_\psi(\mathbf y_{\mathsf s-1}\mid\mathbf y_\mathsf s, \mathbf h_\mathbf c, z).
\end{equation}

\paragraph{Variational distribution.}
We factor the variational posterior as:
\begin{equation}
q(z, \mathbf y_{1:\mathsf S}\mid\mathbf c, \mathbf y_0) = q_\eta(z\mid\mathbf c, \mathbf y_0)\, q(\mathbf y_{1:\mathsf S}\mid\mathbf y_0),
\end{equation}
where $q_\eta(z\mid\mathbf c, \mathbf y_0)$ is a learned categorical encoder and $q(\mathbf y_{1:\mathsf S}\mid\mathbf y_0)=\prod_{\mathsf s=1}^{\mathsf S}q(\mathbf y_\mathsf s\mid\mathbf y_{\mathsf s-1})$ is the fixed forward noising process.

\paragraph{ELBO derivation.}
The marginal log-likelihood is:
\begin{equation}
\log p(\mathbf y_0\mid\mathbf c) = \log\sum_{z}\int p_\theta(z\mid\mathbf c)\, p_\psi(\mathbf y_{0:\mathsf S}\mid\mathbf h_\mathbf c, z)\,d\mathbf y_{1:\mathsf S}.
\end{equation}
Applying the standard variational bound with proposal $q_\eta(z\mid\mathbf c, \mathbf y_0)\,q(\mathbf y_{1:\mathsf S}\mid\mathbf y_0)$:
\begin{align}
\log p(\mathbf y_0\mid\mathbf c)
&\geq \mathbb{E}_{q_\eta(z\mid\mathbf c,\mathbf y_0)\,q(\mathbf y_{1:\mathsf S}\mid\mathbf y_0)}[\log p_\theta(z\mid\mathbf c) + \log p_\psi(\mathbf y_{0:\mathsf S}\mid\mathbf h_\mathbf c, z) \\
&\qquad\qquad\qquad\qquad\quad- \log q_\eta(z\mid\mathbf c,\mathbf y_0) - \log q(\mathbf y_{1:\mathsf S}\mid\mathbf y_0)] \nonumber\\
&= -\mathrm{KL}\!\big(q_\eta(z\mid\mathbf c,\mathbf y_0)\,\|\,p_\theta(z\mid\mathbf c)\big) + \mathbb{E}_{q_\eta(z\mid\mathbf c,\mathbf y_0)}\!\big[\mathcal{L}_{\text{diff}}(\mathbf c, \mathbf y_0; z)\big],
\label{eq:app_elbo}
\end{align}
where the per-mode diffusion term is:
\begin{equation}
\mathcal{L}_{\text{diff}}(\mathbf c, \mathbf y_0; z) := \mathbb{E}_{q(\mathbf y_{1:\mathsf S}\mid\mathbf y_0)}\!\left[\log p_\psi(\mathbf y_{0:\mathsf S}\mid\mathbf h_\mathbf c, z) - \log q(\mathbf y_{1:\mathsf S}\mid\mathbf y_0)\right].
\label{eq:app_Ldiff}
\end{equation}

\paragraph{Diffusion VLB expansion.}
Expanding $\mathcal{L}_{\text{diff}}$ via the standard DDPM variational bound~\citep{mid-14} yields the usual three terms (prior matching, reverse-step KLs, reconstruction), with each reverse step additionally conditioned on $(\mathbf h_{\mathbf c}, z)$.

\paragraph{Discrete enumeration.}
Because $z$ is categorical with $K$ modes, the expectation over $q_\eta(z\mid\mathbf c,\mathbf y_0)$ is computed exactly:
\begin{equation}
\mathbb{E}_{q_\eta(z\mid\mathbf c,\mathbf y_0)}\!\big[\mathcal{L}_{\text{diff}}(\mathbf c, \mathbf y_0; z)\big] = \sum_{k=1}^{K} q_\eta(z{=}k\mid\mathbf c, \mathbf y_0)\,\mathcal{L}_{\text{diff}}(\mathbf c, \mathbf y_0; z{=}k).
\label{eq:app_z_enum}
\end{equation}

\paragraph{Training loss.}
Starting from the ELBO, we introduce a KL weight $\beta$ and minimize the following base objective:
\begin{equation}
\mathcal{J} = -\sum_{k=1}^{K} q_\eta(z{=}k\mid\mathbf c, \mathbf y_0)\,\mathcal{L}_{\text{diff}}(\mathbf c, \mathbf y_0; z{=}k) + \beta\,\mathrm{KL}\!\big(q_\eta(z\mid\mathbf c, \mathbf y_0)\,\|\,p_\theta(z\mid\mathbf c)\big),
\label{eq:app_training_loss}
\end{equation}
where $\beta$ is the KL weight. We augment this objective with a posterior mutual-information reward with coefficient 0.12 and a batch-marginal usage penalty whose coefficient decays from 0.20 to 0.02. The MI reward is $H(\bar q)-\frac{1}{B}\sum_{i=1}^{B}H(q_i)$, where $q_i$ is the posterior mode distribution for sample $i$, $B$ is the batch size, and $\bar q=\frac{1}{B}\sum_i q_i$.
The usage penalty is $\mathrm{KL}(\bar q\|\mathrm{Uniform}(K))$, where $\mathrm{Uniform}(K)$ denotes the uniform distribution over the $K$ modes.

\subsection{Multimodal Backbone Diagnostics}
\label{app:backbone_diagnostics}

\begin{figure}[H]
    \centering
    \includegraphics[width=\linewidth]{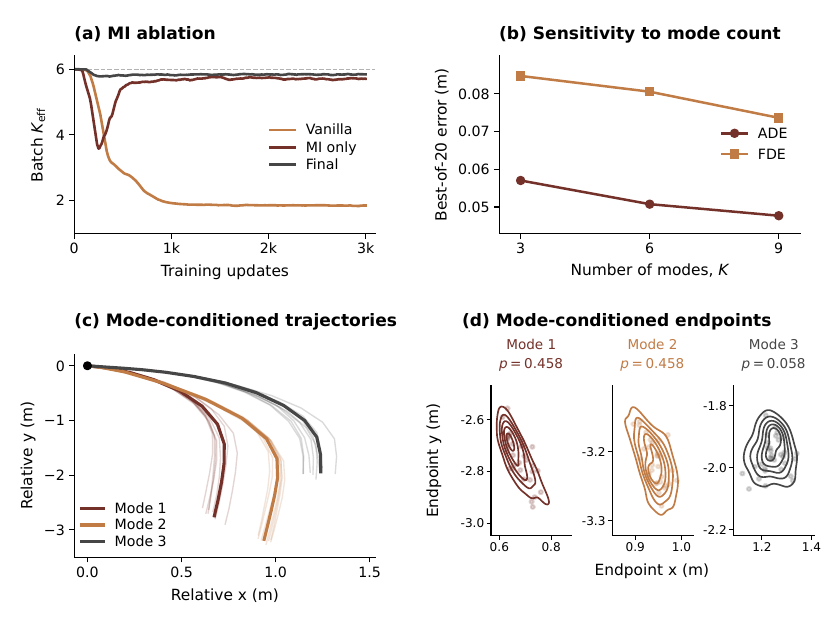}
    \caption{\textbf{Multimodal backbone diagnostics.} (a) Effective mode count during training. Vanilla omits MI and usage regularization, MI only adds the MI reward, and Final includes both. (b) Best-of-20 displacement errors for different mode counts. (c) Three highest-probability modes in one held-out context. Thin lines show eight samples per mode, and thick lines show means over 128 samples. (d) Endpoint density contours estimated from all 128 samples per mode.}
    \label{fig:backbone_diagnostics}
\end{figure}

We assess mode usage, sensitivity to mode count, and conditional distribution shape in the 3-pedestrian Y-corridor setting.

\paragraph{Mode usage.}
We compare matched training configurations on 1,000 held-out contexts.
We measure effective mode usage by posterior perplexity, $K_{\mathrm{eff}}=\exp(H(\bar q))$, analogous to codebook perplexity in Kori et al.~\citep{kori2024grounded}, where $\bar q$ is the evaluation-set average posterior.
A mode is active if it receives the highest posterior probability in at least 1\% of contexts.
Without MI or usage regularization, only two of six modes remain active, with $K_{\mathrm{eff}}=1.84$.
MI alone restores all six modes, with $K_{\mathrm{eff}}=5.79$.
The final backbone also uses all six modes, with $K_{\mathrm{eff}}=5.90$.
Figure~\ref{fig:backbone_diagnostics}(a) shows the corresponding training dynamics.

\paragraph{Sensitivity to mode count.}
We compare $K=3,6,9$ under matched training settings and evaluate best-of-20 prediction accuracy on 1,017 held-out contexts.
ADE/FDE decrease from $0.057/0.085$~m at $K=3$ to $0.048/0.074$~m at $K=9$ (Fig.~\ref{fig:backbone_diagnostics}(b)).
All categories remain active at each tested value of $K$.

\paragraph{Mode-conditioned unimodality.}
We evaluate 200 held-out contexts in the Y-junction decision region.
For each context, we select modes covering at least 95\% of the prior probability and draw 128 trajectories per mode.
We use kernel density estimation in endpoint and trajectory-PCA spaces to identify secondary density basins that persist across bandwidth changes and bootstrap resampling.
Conditionals passing the single-basin diagnostic account for 93.34\% of the evaluated probability mass, with a context-bootstrap 95\% confidence interval of 89.94\% to 96.40\%.
Figures~\ref{fig:backbone_diagnostics}(c) and \ref{fig:backbone_diagnostics}(d) illustrate one context.
Together with the anchor and curvature diagnostics, these results support the local approximation in this setting.

\subsection{Trajectory-Level NLL Calibration}
\label{sec:nll_calibration}

For each pedestrian and latent mode $k\in[K]$, the SCOPE predictor provides a mean trajectory
$\boldsymbol{\mu}_{k}\in\mathbb{R}^{2T}$, a full position covariance
$\boldsymbol{\Sigma}_{k}\in\mathbb{R}^{2T\times 2T}$ over horizon~$T$, and a prior mode weight $\pi_{k}$.
These raw covariances are
not guaranteed to be statistically well calibrated: the off-training
distribution, discretisation of the reverse diffusion process, and
approximations in the precision parametrisation may all bias the predicted
uncertainty. We therefore perform a lightweight post-hoc calibration step on a
disjoint calibration split ($10\%$ of the data, separate from the $70\%$ training and $20\%$ test splits), which learns a small set of inflation factors that
rescale the predicted covariances so that the trajectory-level negative
log-likelihood (NLL) of the ground truth is minimised.

\paragraph{Per-horizon inflation factors.}
We introduce $T$ learnable log-scale parameters
$\mathbf{s}=(s_{1},\dots,s_{T})\in\mathbb{R}^{T}$ and define the per-horizon
inflation factors
\begin{equation}\label{eq:tau_def}
    \tau_{t} \;=\; \exp(s_{t}), \qquad t \in[T],
\end{equation}
so that $\tau_{t}>0$ is automatically enforced and $\tau_{t}=1$ (i.e.\
$s_{t}=0$) corresponds to the uncalibrated predictor. Intuitively,
$\tau_{t}>1$ signals that the SCOPE predictor is \emph{overconfident} at
horizon~$t$ and its covariance should be inflated, while $\tau_{t}<1$
indicates \emph{underconfidence}.

\paragraph{Symmetric scaling of the full covariance.}
Because $\boldsymbol{\Sigma}_{k}$ couples all timesteps through its
off-diagonal blocks, a naive per-timestep scaling of the $2\times 2$ diagonal
blocks would destroy the temporal correlation structure that the precision
head has learned. We instead apply a \emph{symmetric} rescaling that inflates
each timestep's marginal while preserving the correlations:
\begin{equation}\label{eq:D_def}
    \mathbf{D} \;=\; \mathrm{diag}\!\left(
        \sqrt{\tau_{1}},\,\sqrt{\tau_{1}},\;
        \sqrt{\tau_{2}},\,\sqrt{\tau_{2}},\;
        \dots,\;
        \sqrt{\tau_{T}},\,\sqrt{\tau_{T}}
    \right)\in\mathbb{R}^{2T\times 2T},
\end{equation}
where each $\sqrt{\tau_{t}}$ is repeated twice to act identically on the
$(x,y)$ coordinates of timestep~$t$. The calibrated full covariance is then
\begin{equation}\label{eq:sigma_cal}
    \boldsymbol{\Sigma}_{k}^{\text{cal}}
    = \mathbf{D}\boldsymbol{\Sigma}_{k}\mathbf{D}.
\end{equation}
Since $\mathbf{D}$ is diagonal with positive entries, $\mathbf{D}$ is its own
transpose and $\boldsymbol{\Sigma}_{k}^{\text{cal}}$ remains symmetric
positive-definite whenever $\boldsymbol{\Sigma}_{k}$ is, so the Gaussian
density is always well defined. The diagonal blocks of
$\boldsymbol{\Sigma}_{k}^{\text{cal}}$ become $\tau_{t}\,\boldsymbol{\Sigma}_{k,tt}$,
recovering per-timestep inflation as a special case, while the off-diagonal
blocks are rescaled by $\sqrt{\tau_{s}\tau_{t}}$, preserving the correlation
structure.

\paragraph{Trajectory-level GMM log-likelihood.}
Let $\mathbf{y}\in\mathbb{R}^{2T}$ denote the flattened ground-truth
trajectory of a pedestrian. The log-density of $\mathbf{y}$ under mode~$k$
with the calibrated covariance is
\begin{equation}\label{eq:logN}
    \log\mathcal{N}\!\bigl(
        \mathbf{y}\,;\,\boldsymbol{\mu}_{k},\,
        \mathbf{D}\boldsymbol{\Sigma}_{k}\mathbf{D}
    \bigr)
    \;=\; -\tfrac{1}{2}\Bigl[
        2T\log(2\pi)
        + \log\bigl|\mathbf{D}\boldsymbol{\Sigma}_{k}\mathbf{D}\bigr|
        + \Delta_{k}^{\!\top}\bigl(\mathbf{D}\boldsymbol{\Sigma}_{k}\mathbf{D}\bigr)^{-1}\Delta_{k}
    \Bigr],
\end{equation}
with residual $\Delta_{k}=\mathbf{y}-\boldsymbol{\mu}_{k}$. Rather than
forming $\mathbf{D}\boldsymbol{\Sigma}_{k}\mathbf{D}$ explicitly, we exploit
the diagonal structure of $\mathbf{D}$ to obtain
\begin{align}
    \log\bigl|\mathbf{D}\boldsymbol{\Sigma}_{k}\mathbf{D}\bigr|
    &\;=\; 2\sum_{t=1}^{T}\log\tau_{t}
        \;+\; \log\bigl|\boldsymbol{\Sigma}_{k}\bigr|,
    \label{eq:logdet}\\[2pt]
    \Delta_{k}^{\!\top}\bigl(\mathbf{D}\boldsymbol{\Sigma}_{k}\mathbf{D}\bigr)^{-1}\Delta_{k}
    &\;=\; \bigl(\mathbf{D}^{-1}\Delta_{k}\bigr)^{\!\top}
           \boldsymbol{\Sigma}_{k}^{-1}
           \bigl(\mathbf{D}^{-1}\Delta_{k}\bigr).
    \label{eq:mahal}
\end{align}
Equations~\eqref{eq:logdet}--\eqref{eq:mahal} are implemented with a single
Cholesky factorisation $\boldsymbol{\Sigma}_{k}=\mathbf{L}_{k}\mathbf{L}_{k}^{\!\top}$
for each sample and mode, yielding
$\log|\boldsymbol{\Sigma}_{k}|=2\sum_{i}\log\mathbf{L}_{k,ii}$ and a
triangular solve for the Mahalanobis term. A small diagonal jitter
$\epsilon\,\mathbf{I}$ is added to $\boldsymbol{\Sigma}_{k}$ before
factorisation for numerical stability.

The log-likelihood of the full trajectory under the Gaussian mixture is
obtained by marginalising over modes \emph{at the trajectory level}:
\begin{equation}\label{eq:traj_gmm}
    \log p(\mathbf{y})
    \;=\; \operatorname*{logsumexp}_{k\in[K]}\!
    \Bigl[\,
        \log\pi_{k}
        \;+\; \log\mathcal{N}\!\bigl(
            \mathbf{y}\,\big|\,\boldsymbol{\mu}_{k},\,
            \mathbf{D}\boldsymbol{\Sigma}_{k}\mathbf{D}
        \bigr)
    \,\Bigr].
\end{equation}
Crucially, the \texttt{logsumexp} is applied once over the entire horizon,
not inside an outer sum over timesteps. A per-timestep formulation
$\sum_{t}\mathrm{logsumexp}_{k}[\log\pi_{k}+\log\mathcal{N}(y_{t}\mid\mu_{k,t},\tau_{t}\Sigma_{k,tt})]$
implicitly permits the predicted mode to switch between consecutive timesteps. It therefore does not preserve the shared trajectory-level mode or the within-mode temporal
correlations represented by the full covariance $\Sigma_k$. The trajectory-level form in
Eq.~\eqref{eq:traj_gmm} is the correct joint log-likelihood of $\mathbf{y}$
under the Gaussian mixture and is the quantity we calibrate against.

\paragraph{Calibration objective.}
Given a calibration set $\mathcal{D}_{\text{cal}}=\{\mathbf{y}^{(b)}\}_{b=1}^{B}$
of ground-truth trajectories together with their predicted GMM parameters
$\{\pi_{k}^{(b)},\boldsymbol{\mu}_{k}^{(b)},\boldsymbol{\Sigma}_{k}^{(b)}\}_{k=1}^{K}$,
we minimise the mean trajectory-level NLL with an $\ell_{2}$ regulariser on
the log-scales:
\begin{equation}\label{eq:nll_obj}
    \mathcal{L}(\mathbf{s})
    \;=\; -\,\frac{1}{B}\sum_{b=1}^{B}
          \log p\!\left(\mathbf{y}^{(b)};\,\mathbf{s}\right)
          \;+\; \lambda_{\text{reg}}\sum_{t=1}^{T}s_{t}^{2}.
\end{equation}
The regulariser $\lambda_{\text{reg}}\sum_{t}s_{t}^{2}$ shrinks each
$\tau_{t}$ toward~$1$, discouraging extreme inflation factors that could
overfit the calibration split, and is equivalent to a zero-mean Gaussian
prior on the log-scales. Only the $T$ parameters in $\mathbf{s}$ are updated;
the predicted mode weights $\pi_{k}$, means $\boldsymbol{\mu}_{k}$, and raw covariances
$\boldsymbol{\Sigma}_{k}$ remain frozen.

\paragraph{Optimisation and data split.}
The overall dataset is partitioned into training ($70\%$), test ($20\%$), and calibration ($10\%$) splits. We collect GMM predictions on the calibration split to fit~$\mathbf{s}$. All reported metrics are computed exclusively on the disjoint test split.
Equation~\eqref{eq:nll_obj} is minimised with Adam for a fixed number of
iterations, with all mode parameters treated as constants. The only tensors
that receive gradients are the $T$ entries of~$\mathbf{s}$. At inference,
calibration is applied by rescaling every predicted full covariance via
Eq.~\eqref{eq:sigma_cal}, after which the calibrated GMM can be used
interchangeably with the uncalibrated one.

\subsection{Hutchinson Trace Estimator: From Matrix Matching to Probe Matching}\label{app:hutchinson}

We train the SCOPE predictor to produce a precision matrix $\boldsymbol\Lambda_\phi(\mathbf y)$ that approximates the teacher curvature $\boldsymbol\Lambda^*(\mathbf y):=-\nabla_{\mathbf y}^2\log p(\mathbf y)$. The most direct objective is the Frobenius-norm error:
\begin{equation}\label{eq:frob_obj}
\min_\phi\;\|\boldsymbol\Lambda_\phi(\mathbf y)-\boldsymbol\Lambda^*(\mathbf y)\|_F^2.
\end{equation}
However, $\boldsymbol\Lambda^*(\mathbf y)\in\mathbb{R}^{dT\times dT}$ is a dense matrix that is prohibitively expensive to compute, store, and supervise element-wise. We therefore seek an equivalent objective that avoids materializing the full matrix.

\paragraph{Hutchinson identity.}
Let $\mathbf M:=\boldsymbol\Lambda_\phi(\mathbf y)-\boldsymbol\Lambda^*(\mathbf y)$ denote the error matrix. Then:
\begin{equation}
\|\mathbf M\|_F^2 = \mathrm{tr}(\mathbf M^\top \mathbf M).
\end{equation}
For any random vector $\mathbf u$ satisfying the isotropic condition $\mathbb{E}[\mathbf u\mathbf u^\top]=\mathbf I$ (e.g.\ $\mathbf u\sim\mathcal{N}(\mathbf 0,\mathbf I)$), the Hutchinson trace estimator gives:
\begin{align}
\mathbb{E}_{\mathbf u}\!\left[\|\mathbf M\mathbf u\|_2^2\right]
&= \mathbb{E}_{\mathbf u}\!\left[\mathbf u^\top \mathbf M^\top \mathbf M\,\mathbf u\right] \nonumber\\
&= \mathrm{tr}\!\left(\mathbf M^\top \mathbf M\;\mathbb{E}[\mathbf u\mathbf u^\top]\right) \nonumber\\
&= \mathrm{tr}(\mathbf M^\top \mathbf M) \nonumber\\
&= \|\mathbf M\|_F^2.
\label{eq:hutchinson_frob}
\end{align}
Therefore, the Frobenius-norm objective decomposes as:
\begin{equation}\label{eq:frob_to_probe}
\|\boldsymbol\Lambda_\phi(\mathbf y)-\boldsymbol\Lambda^*(\mathbf y)\|_F^2
\;=\;
\mathbb{E}_{\mathbf u}\!\left[\|(\boldsymbol\Lambda_\phi(\mathbf y)-\boldsymbol\Lambda^*(\mathbf y))\mathbf u\|_2^2\right].
\end{equation}
This means we do not need to compare the full matrices directly; it suffices to compare how they act on random probe directions $\mathbf u$. Matching the operator response on enough random probes is equivalent, in expectation, to matching the matrices in Frobenius norm. This replaces full-matrix comparison with $dT$-dimensional probe responses and avoids materializing the teacher matrix.

\subsection{HVP Identity: Efficient Teacher Signal}\label{app:hvp_identity}

The Hutchinson identity (Appendix~\ref{app:hutchinson}) specifies the target: the operator response on random probes. We still need an efficient way to compute the teacher's probe response $\boldsymbol\Lambda^*(\mathbf y)\mathbf u$. This is where the Hessian--vector product (HVP) identity enters.

\paragraph{Score--curvature relationship.}
Recall that the teacher curvature is the negative Hessian of the terminal log-density:
\begin{equation}
\boldsymbol\Lambda^*(\mathbf y) = -\nabla_{\mathbf y}^2 \log p(\mathbf y),
\end{equation}
The score model approximates this gradient:
\begin{equation}
s_\theta(\mathbf y, 0) \approx \nabla_{\mathbf y}\log p(\mathbf y).
\end{equation}
In implementation, we approximate the terminal score using a small nonzero diffusion noise level.

\paragraph{HVP derivation.}
For any probe vector $\mathbf u\in\mathbb{R}^{dT}$, the teacher's action on $\mathbf u$ can be written as:
\begin{align}
\boldsymbol\Lambda^*(\mathbf y)\,\mathbf u
&= \left[-\nabla_{\mathbf y}^2 \log p(\mathbf y)\right]\mathbf u \nonumber\\
&= -\nabla_{\mathbf y}\!\left[\nabla_{\mathbf y}\log p(\mathbf y)^\top \mathbf u\right] \nonumber\\
&\approx -\nabla_{\mathbf y}\!\left(s_\theta(\mathbf y, 0)^\top \mathbf u\right).
\label{eq:hvp_derivation}
\end{align}
The key observation is that $s_\theta(\mathbf y,0)^\top \mathbf u$ is a \emph{scalar}, so its gradient with respect to $\mathbf y$ is a vector of the same dimension as $\mathbf y$. This is precisely a Hessian--vector product (HVP), computable via a single backward pass through automatic differentiation \emph{without} materializing the $dT\times dT$ Hessian.

\paragraph{Final training loss.}
Using the score-based approximation to the teacher response yields the following probe-matching loss:
\begin{equation}
\mathcal{L}_{\text{prec}}(\phi)
\;=\;
\mathbb{E}_{\mathbf y}\,\mathbb{E}_{\mathbf u}\!\left[\left\|\boldsymbol\Lambda_\phi(\mathbf y)\mathbf u + \nabla_{\mathbf y}\!\left(s_\theta(\mathbf y,0)^\top \mathbf u\right)\right\|_2^2\right].
\label{eq:prec_loss_final}
\end{equation}
This converts expensive matrix-level supervision into efficient vector-level supervision. In summary, Hutchinson tells us \emph{why} matching probe responses suffices (it is equivalent to Frobenius-norm matching), and HVP tells us \emph{how} to compute the teacher's probe response efficiently (via a single autodiff backward pass through the score model).

\section{MPPI Control Details}\label{Section: MPPI Control}

\subsection{MPPI Cost Function Design}
\label{app:mppi_cost}

We employ Model Predictive Path Integral (MPPI) control~\cite{williams2017mppi} to compute
real-time navigation commands. At each planning step, the controller draws
$N$ action sequences $\{\mathbf{u}^{(k)}_{0:H-1}\}_{k=1}^{N}$, rolls them forward
through a deterministic dynamics model to obtain state trajectories
$\{\mathbf{x}^{(k)}_{0:H}\}_{k=1}^{N}$, evaluates a stage cost on each, and returns the
importance-weighted average as the optimal action. The total trajectory cost
for a sample $k\in[N]$ is
\begin{equation}\label{eq:mppi_total_cost}
    S^{(k)} \;=\; w_{\text{goal}}\, J_{\text{goal}}^{(k)}
             \;+\; w_{\text{wall}}\, J_{\text{wall}}^{(k)}
             \;+\; w_{\text{gmm}}\, J_{\text{gmm}}^{(k)},
\end{equation}
where the three active terms are detailed below.

\subsubsection{Goal-Reaching Cost}

The goal cost encourages the robot to approach the target position
$\mathbf{p}_{\text{goal}}\in\mathbb{R}^{2}$ at every timestep along the
planning horizon~$H$:
\begin{equation}\label{eq:goal_cost}
    J_{\text{goal}} = \sum_{t=1}^{H}
    \bigl\lVert \mathbf{p}_{t} - \mathbf{p}_{\text{goal}} \bigr\rVert_{2},
\end{equation}
where $\mathbf{p}_{t}\in\mathbb{R}^{2}$ is the robot position at planning step~$t$.
We sum the Euclidean distance over the full horizon rather than penalising
only the terminal state. This produces a smooth gradient that guides intermediate
waypoints toward the goal.

\subsubsection{Wall-Avoidance Cost}

Navigation occurs inside a Y-junction corridor whose walls are described by a
geometry module that returns, for any query point, the minimum clearance
$c_{t}$ to the nearest wall segment. To avoid hard discontinuities in the cost
landscape, we apply a soft barrier based on the \emph{softplus} function
$\operatorname{softplus}(x) = \ln(1+e^{x})$:
\begin{equation}\label{eq:wall_cost}
    J_{\text{wall}} = \sum_{t=1}^{H}
    \operatorname{softplus}\!\left(
        \frac{d_{\text{wall}} - c_{t}}{\epsilon}
    \right),
\end{equation}
where $d_{\text{wall}}$ is a comfortable clearance margin and $\epsilon$ is a
smoothing parameter that controls the barrier sharpness. When the robot
clearance $c_t$ exceeds $d_{\text{wall}}$, the argument is negative and the
cost decays exponentially. As $c_t$ falls below $d_{\text{wall}}$, the cost
rises steeply, yielding a smooth yet aggressive repulsive field near walls.

\subsubsection{GMM Chance-Constrained Collision Cost}
\label{sec:gmm_chance}

Rather than inflating obstacle radii by a fixed safety margin, we formulate
collision avoidance as a \emph{soft chance constraint} over the multimodal
predictions supplied by the forecaster. Each agent $n$ is described by a
Gaussian Mixture Model (GMM) with $K$ modes, where mode $k\in[K]$ carries normalized probability weight
$\pi_{k}^{(n)}$, mean trajectory $\boldsymbol{\mu}_{k,t}^{(n)}$, and time-varying
covariance $\boldsymbol{\Sigma}_{k,t}^{(n)}$.

\textbf{Chance constraint formulation. }
We denote $C_n(t)$ to be the event that a collision happens between the robot and the $n$-th agent at time $t$. By modeling the robot and the $n$-th agent as disk $\calA(\p_t,r_r)$ and $\calA(\p_t^{(n)},r_h)$ respectively, the collision event between these two dynamic objects at time $t$ is
\begin{equation}
    C_n(t):=\{\omega\in\Omega:\|\p_t^{(n)}(\omega)-\p_t\|\le r\},
\end{equation}
where $r=r_r+r_h$.
With $N_p$ agents, the collision event between the robot and any of the agents at time $t$ is
\begin{equation}
    C(t):=\bigcup_{n=1}^{N_p}C_n(t).
\end{equation}
Chance constraints typically upper bound the collision probability $\Pr(C(t))\le\delta$, with a positive threshold $\delta>0$. However, calculating $\Pr(C(t))$ is not trivial. Instead, we chose a simpler and more conservative approach by calculating its union bound (Boole's inequality) as follows:
\begin{equation}
\Pr(C(t))=\Pr\left(\bigcup_{n=1}^{N_p}C_n(t)\right)\le\sum_{n=1}^{N_p}\Pr(C_n(t)).
\end{equation}
From here, we can utilize the derivation results from \ref{app:cc_gmm} and approximate $\Pr(C_n(t))$ to be
\begin{equation}
    \Pr(C_n(t))\le\sum_{k=1}^{K}{\pi_k^{(n)}\cdot\Phi\left(\frac{r-d_{k,t}^{(n)}}{\sigma_{k,t}^{(n)}}\right)}.
\end{equation}
Denote the union bound risk to be
\begin{equation}
P_\text{risk}(t):=\sum_{n=1}^{N_p}\sum_{k=1}^{K}\pi_k^{(n)} \cdot \Phi\left(\frac{r-d_{k,t}^{(n)}}{\sigma_{k,t}^{(n)}}\right).
\end{equation}
If $P_\text{risk}(t)\le\delta$, the collision probability under the predictive GMM is also bounded by $\delta$:
\begin{equation}
    \Pr(C(t))\le P_\text{risk}(t)\le\delta.
\end{equation}

\textbf{Soft penalty. }
We penalize violations of the per-timestep risk budget using a squared softplus penalty:
\begin{equation}\label{eq:gmm_cost}
    J_{\text{gmm}}
    = \sum_{t=1}^{H}
    \left[
        \operatorname{softplus}\!\left(
            \frac{P_{\text{risk}}(t) - \delta}{\tau}
        \right)
    \right]^{2},
\end{equation}
where $\tau$ is a temperature that controls penalty sharpness. The squared softplus provides a smooth, twice-differentiable cost that is effectively zero when $P_{\text{risk}}(t) < \delta$ yet grows quadratically once the risk
budget is exceeded, encouraging the planner to redistribute probability mass
toward safer trajectories.

\subsection{Derivation of the Collision Probability Upper Bound (one agent)}\label{app:cc_gmm}
Here, we demonstrate how to derive the collision probability upper bound. Recall that given the $n$-th agent ($n\in[N_p]$), time $t$ and the per-time context $\mathbf c_t^{(n)}$, we have
\begin{equation}
\mathbf{p}_t^{(n)}\sim\sum_{k=1}^{K}\pi_k^{(n)}\mathcal{N}(\boldsymbol\mu_{k,t}^{(n)},\boldsymbol{\Sigma}_{k,t}^{(n)}).
\end{equation}
Define the collision area to be
\begin{equation}
    C:=\{\mathbf x\in\mathbb R^2:\|\mathbf x-\mathbf p_t\|\le r\},
\end{equation}
where $\mathbf p_t$ is the robot's position at time $t$ and $r=r_h+r_r$ is the sum of the human and robot radii.
For each mode $k\in[K]$, define
\begin{equation}
\boldsymbol\nu_{k,t}^{(n)}:=\boldsymbol\mu_{k,t}^{(n)} - \mathbf p_t, \label{def:nu}
\end{equation}
to be the displacement and its unit direction $\hat{\boldsymbol\nu}$.
Define
\begin{equation}
    \mathcal H:=\{\mathbf x\in\mathbb R^2:\hat{\boldsymbol\nu}^\top(\mathbf x - \mathbf p_t)\le r\}
\end{equation}
to be the half space induced by $\hat{\boldsymbol\nu},\ \mathbf p_t,\ r$. Intuitively, this is the half-space on the robot's side, opposite the human's mode position.
It's clear that $C\subseteq \mathcal H$ because if $\mathbf x\in C$, then by the Cauchy-Schwarz inequality, one has
$\hat{\boldsymbol\nu}^\top(\mathbf x - \mathbf p_t)
\le\|\mathbf x - \mathbf p_t\|
\le r$.
Hence, the following holds:
\begin{equation}
\Pr(\mathbf x\in C)\le\Pr(\mathbf x\in \mathcal H).
\end{equation}
Let $\mathbf x_k:=(\mathbf p_t^{(n)}|z=k) \sim \mathcal N(\boldsymbol\mu_{k,t}^{(n)},\boldsymbol\Sigma_{k,t}^{(n)})$. Centering $\mathbf x_k$ at $\mathbf p_t$, we obtain
\begin{equation}
    \mathbf x_k-\mathbf p_t \sim \mathcal N(\boldsymbol\nu_{k,t}^{(n)},\boldsymbol\Sigma_{k,t}^{(n)}).
\end{equation}
Left-multiplying $\mathbf x_k-\mathbf p_t$ by $\hat{\boldsymbol\nu}^\top$ yields
\begin{equation}
    \hat{\boldsymbol\nu}^\top(\mathbf x_k-\mathbf p_t) \sim \mathcal N(d_{k,t}^{(n)},(\sigma_{k,t}^{(n)})^2),
\end{equation}
where $d_{k,t}^{(n)}=\hat{\boldsymbol\nu}^\top\boldsymbol\nu_{k,t}^{(n)}$ and  $(\sigma_{k,t}^{(n)})^2=\hat{\boldsymbol\nu}^\top\boldsymbol\Sigma_{k,t}^{(n)}\hat{\boldsymbol\nu}$. Then we have
\begin{align}
    \Pr(\mathbf x_k\in\mathcal H)
    &=\Pr(\hat{\boldsymbol\nu}^\top(\mathbf x_k-\mathbf p_t)\le r)\\
    &=\Pr\left(\frac{\hat{\boldsymbol\nu}^\top(\mathbf x_k-\mathbf p_t)-d_{k,t}^{(n)}}{\sigma_{k,t}^{(n)}}\le \frac{r-d_{k,t}^{(n)}}{\sigma_{k,t}^{(n)}}\right)\\
    &=\Phi\left(\frac{r-d_{k,t}^{(n)}}{\sigma_{k,t}^{(n)}}\right)
\end{align}
Hence, we have the collision probability (one agent) upper bounded as follows:
\begin{align}
    \Pr(C_n(t))
    &=\Pr(\mathbf p_t^{(n)}\in C)\\
    &=\sum_{k=1}^{K}\Pr(z=k)\Pr(\mathbf p_t^{(n)}\in C \mid z=k) \\
    &=\sum_{k=1}^{K}\pi_k^{(n)}\Pr(\mathbf x_k\in C) \\
    &\le\sum_{k=1}^{K}\pi_k^{(n)}\Pr(\mathbf x_k\in \mathcal H) \\
    &=\sum_{k=1}^{K}\pi_k^{(n)}\cdot\Phi\left(\frac{r-d_{k,t}^{(n)}}{\sigma_{k,t}^{(n)}}\right).
\end{align}

\subsection{\mymethod{} Algorithm Details}
\label{sec:appendix:scope-algs}

\subsubsection{Notation and overview}
\label{sec:appendix:scope-algs:notation}

Let $p_{0:L-1} \in \mathbb{R}^{L\times 2}$ denote the global path ($L$ waypoints after resampling), with cumulative arclength $s_i = \sum_{k=0}^{i-1} \|p_{k+1} - p_{k}\|_{2}$ precomputed once per episode. The robot has state $x_t$, action $u_t$, and single-integrator dynamics $x_{t+1} = x_t + u_t\,\Delta t$. For every waypoint the SCOPE predictor predicts a per-timestep state covariance $\Sigma^x_i$ and precision $\Lambda^x_i = (\Sigma^x_i)^{-1}$, queried once per episode.

\subsubsection{Nominal MPPI control sequence}
\label{sec:appendix:scope-algs:nominal}

The MPPI inner loop is standard receding-horizon shooting, but the
nominal control sequence $U \in \mathbb{R}^{H \times 2}$ is warm-started
from the diffusion path so that even the first set of $N$ rollouts is
centered on the diffusion-generated reference trajectory.

\paragraph{Warm-start from diffusion path.}
At episode start the nominal sequence is initialized from path velocities:
\begin{equation}
  U^{(0)}_t \;=\; \frac{p_{t+1} - p_{t}}{\Delta t},
  \qquad t = 0, \ldots, H-2,
  \label{eq:appendix:scope-algs:warmstart}
\end{equation}
with $U^{(0)}_{H-1} = U^{(0)}_{H-2}$ (zero-acceleration extrapolation). Because the diffusion prior encodes the maze geometry, this centers initial rollouts on the diffusion-generated reference trajectory rather than near-random control sequences.

\paragraph{Receding-horizon refinement.}
At each subsequent step the controller shifts the nominal forward by one timestep, draws $N$ perturbations $\delta u^{(k)}_t \sim \mathcal{N}(0,\,\Sigma^{u}_t)$ using the tube covariance from Sec.~\ref{sec:appendix:scope-algs:tube}, rolls them out, and updates the nominal via the standard MPPI~\cite{williams2017mppi} softmax-weighted average. The running cost combines a goal-tracking term toward the subgoal $g_t$ from Sec.~\ref{sec:appendix:scope-algs:subgoal} with an obstacle collision penalty around each wall.

\subsubsection{Subgoal selection}
\label{sec:appendix:scope-algs:subgoal}

Subgoal selection produces the target position $g_t$ for the MPPI running cost via three components.

\paragraph{Monotone nearest index.}
Naive nearest-waypoint projection can snap backwards when the path self-intersects. We restrict the search to a forward-only window of $W = 20$ waypoints:
\begin{equation}
  i_t \;=\; \arg\min_{j \in [\,i_{t-1},\ \min(i_{t-1}+W,\ L-1)\,]}\; \|p_j - x_t\|_{2},
  \qquad
  i_0=0.
  \label{eq:appendix:scope-algs:monotone}
\end{equation}

\paragraph{Arclength-based lookahead.}
The subgoal is the first waypoint whose arclength exceeds the current by a lookahead distance $L_{\text{dist}}$:
\begin{equation}
  i^{\text{sg}}_t \;=\; \min\!\left\{\,j \ge i_t \;:\; s_j \;\ge\; s_{i_t} + L_{\text{dist}}\,\right\}.
  \label{eq:appendix:scope-algs:arclength}
\end{equation}
When the remaining path is shorter than the lookahead, or the path index is within a few waypoints of the end, the subgoal switches to the true goal~$g$.

\subsubsection{Precision tubes for exploration covariance}
\label{sec:appendix:scope-algs:tube}

The SCOPE predictor provides per-waypoint state covariances $\{\Sigma^x_i\}_{i=0}^{L-1}$ along the global path. These are converted into the time-varying action-sampling covariance $\Sigma^{u}_t$ for MPPI via the pipeline in Algorithm~\ref{alg:appendix:scope-algs:pipeline}. The conversion runs once per episode; at each control step MPPI retrieves the appropriate covariance by index lookup.

\paragraph{State-to-action mapping.}
At episode start the SCOPE predictor is queried on the path:
\begin{equation}
  \{\Sigma^x_i\}_{i=0}^{L-1} \;=\; \mathrm{SCOPE}\!\big(p_{0:L-1},\ z_{\text{wall}}\big),
  \label{eq:appendix:scope-algs:phquery}
\end{equation}
and the marginal covariances are rescaled to unnormalized configuration space. We scale the next-waypoint covariance to set the action exploration covariance under single-integrator dynamics:
\begin{equation}
  \Sigma^u_i \;=\; \frac{\gamma^{2}}{\Delta t^{2}}\;\Sigma^x_{i+1},
  \label{eq:appendix:scope-algs:tube2action}
\end{equation}
where $\gamma < 1$ is a tube-radius scale factor. Each $\Sigma^u_i$ is then eigendecomposed.
All covariance windows of length $H$ along the path are precomputed once; at control step~$t$ the covariance is retrieved by a single lookup indexed by the current path position~$i_t$.

\begin{algorithm}[ht]
\caption{Precision tube to MPPI exploration covariance.}
\label{alg:appendix:scope-algs:pipeline}
\begin{algorithmic}[1]
  \Statex \textbf{Per episode} (after the diffusion path $p_{0:L-1}$ is selected):
  \State $\{\Sigma^x_i\} \gets \mathrm{SCOPE}(p_{0:L-1},\,z_{\text{wall}})$; rescale to configuration space
  \State $\Sigma^u_i \gets (\gamma^{2}/\Delta t^{2})\,\Sigma^x_{i+1}$ for $i=0,\ldots,L-2$ \Comment{Eq.~\eqref{eq:appendix:scope-algs:tube2action}}
  \State Precompute Cholesky factors for sliding windows of length $H$
  \Statex \textbf{Per control step} $t$:
  \State Retrieve the cached covariance window at path index $i_t$.
  \State Sample $\delta u^{(k)}_t \sim \mathcal{N}(0,\,\Sigma^{u}_t)$ via cached Cholesky factor
\end{algorithmic}
\end{algorithm}

\section{Experiment Details}\label{app:experiment_details}

\begin{table}[ht]
    \centering
    \footnotesize
    \caption{
        Trajectory accuracy (ADE\,/\,FDE). All \mymethod{} variants share the same backbone and achieve identical ADE/FDE.
        \textbf{Bold}: best.
    }
    \label{tab:ph_ade_fde}
    \setlength{\tabcolsep}{5pt}
    \begin{tabular}{l|cccc}
        \toprule
        \textbf{Method} & $N_p{=}3$ & $N_p{=}6$ & $N_p{=}9$ & $N_p{=}12$ \\
        \midrule
          LKF          & 0.55/1.22 & 0.48/1.03 & 0.41/0.92 & 0.30/0.67 \\
          MID-Ens      & 0.07/0.12 & 0.12/0.23 & 0.10/0.21 & 0.12/0.24 \\
          \midrule
          Ours         & \textbf{0.07}/\textbf{0.11} & \textbf{0.10}/\textbf{0.18} & \textbf{0.10}/\textbf{0.20} &
  \textbf{0.12}/\textbf{0.24} \\
        \bottomrule
    \end{tabular}
\end{table}

This section provides implementation details for three experiments:
forecaster quality on the Y-corridor (Q1, Sec.~\ref{exp:ph_property}),
closed-loop Y-corridor navigation with MPPI (Q3, Sec.~\ref{exp:y_corridor}),
and MPPI sample efficiency on Maze2D (Q2, Sec.~\ref{exp:maze2d}).
Q1 and Q3 share the same MID checkpoints and Precision Tube Head, differing
only in evaluation: Q1 replays ground-truth trajectories, Q3 generates them
online with ORCA and feeds the GMM into an MPPI controller. Q2 uses an
independent diffusion prior and Precision Tube Head trained on Maze2D.

\subsection{Shared Y-Corridor Setup}\label{app:exp_shared}

\textbf{MID backbone.}
Histories are encoded by the Trajectron++~\citep{salzmann2021trajectrondynamicallyfeasibletrajectoryforecasting} encoder (per-pedestrian temporal RNN with social-attention), producing a $256$-dimensional context vector per pedestrian. A discrete latent $z\in\{1,\dots,K_z\}$ with $K_z=6$ modes is attached. The backbone training objective is described in Appendix~\ref{app:mixture_training}. Conditioned on $(c, z)$, the diffusion decoder uses $\epsilon$-prediction with a $100$-step linear variance schedule. At inference, we use $5$ DDIM~\citep{song2022denoisingdiffusionimplicitmodels} steps instead of $100$ DDPM steps.

\textbf{Precision Tube Head.}
With the MID backbone frozen, the SCOPE predictor produces a block lower-banded factor $\mathbf L$ and a low-rank factor $\mathbf R$, both FiLM-conditioned on $(c,z)$. Together, they define the trajectory precision $\boldsymbol\Lambda=\mathbf L\mathbf L^\top+\mathbf R\mathbf R^\top$. It is supervised by the HVP self-consistency loss (Appendices~\ref{app:hutchinson} and \ref{app:hvp_identity}).

\textbf{Calibration.}
Trajectory-level $\tau_t$ calibration (Sec.~\ref{sec:nll_calibration}) is applied: the $T$ log-scales $\mathbf{s}$ are fitted via Eq.~\eqref{eq:nll_obj} with Adam on the dedicated calibration split ($10\%$ of the data). All reported metrics use the disjoint test split ($20\%$).

\subsection{Forecaster Quality on Y-Corridor}\label{app:exp_pth}

\textbf{Environments and datasets.}

The Y-corridor is a junction environment in which pedestrians traverse
between three corridor mouths and interact at the centre, exposing
predictors to crossing, merging, and yielding behaviours. We use four
datasets, one per crowd density $N_p\in\{3,6,9,12\}$, generated by
simulating ORCA pedestrians with randomised starts and goals over the three
mouths. All datasets run at $\Delta t=0.4$\,s. For each pedestrian we
observe a history of $H_{\text{obs}}=8$ steps ($3.2$\,s) and predict
$T=12$ steps ($4.8$\,s), matching the Trajectron++/MID convention.

\textbf{Evaluation protocol.}

For each dataset we sample prediction--ground-truth pairs from the test split, collecting up to $1{,}000$ samples per dataset with capping per scene to ensure scenario diversity. The trajectory-level NLL is the joint NLL of the full $T$-step
ground-truth trajectory under the predicted GMM
(Sec.~\ref{sec:nll_calibration}, Eq.~\eqref{eq:traj_gmm}); it does not allow
the predicted mode to switch between timesteps. Best-of-$K$ ADE and FDE are minimized independently over the predicted mean trajectories. To reduce variance
from unbalanced scene sampling, all metrics are aggregated at the
sequence level: per-instance metrics are first averaged within each scene,
and we report the mean $\pm$ sample standard deviation across scenes.

\textbf{Baselines.}

We compare three forecasters on each dataset.
\begin{itemize}[leftmargin=*]
    \item \textbf{LKF} is a linear Kalman filter with constant-velocity
          dynamics on $[p_x, p_y, v_x, v_y]^{\!\top}$. The filter consumes
          the full $H_{\text{obs}}$ observations through predict-update
          cycles and propagates the state forward for $T$ steps; the
          propagated covariance is the predictive uncertainty. LKF outputs
          a single Gaussian ($K{=}1$); noise parameters were tuned on
          the training split.
    \item \textbf{MID-Ens} is an empirical-GMM baseline. We draw $M\in\{64,128,256\}$
          trajectory samples from the MID diffusion model and fit a
          $K{=}3$-component GMM with full covariance in the flattened trajectory space
          $\mathbb{R}^{2T}$. MID-Ens uses exactly
          the same MID model as Ours but replaces the precision head with
          empirical sample statistics, which isolates the contribution of
          \mymethod{}.
    \item \textbf{Ours} (Banded+LR) is the proposed predictor. The MID
          prior $p(z\mid c)$ assigns weights to $K_z=6$ discrete modes; we
          keep the top-$K{=}3$ modes by probability and renormalise. For
          each retained mode we extract the mean trajectory from a single
          DDIM pass and the full $2T\times 2T$ position covariance from
          the Precision Tube Head, then apply per-horizon trajectory-level
          $\tau_t$ calibration (Sec.~\ref{sec:nll_calibration}). Mode
          weights $\pi_k$ are not calibrated. The structural ablations
          \textbf{Diag}, \textbf{Banded}, and \textbf{Banded+LR (no~cal.)}
          share this pipeline and differ only in how the Precision Tube
          Head parameterises uncertainty (independent per-timestep
          variances; Markov band only; full Markov band plus low-rank tail
          without post-hoc calibration, respectively).
\end{itemize}

\textbf{Implementation.}
The shared MID backbone, Precision Tube Head, and calibration procedure are described in Sec.~\ref{app:exp_shared}.

\begin{table}[ht]
\caption{Hyperparameters for Y-corridor experiments (Q1 and Q3). Controller and cost settings apply to Q3 only; all other settings are shared.}
\label{tab:exp_pth_hp}
\centering
\small
\begin{tabular}{llr}
\toprule
\textbf{Category} & \textbf{Setting} & \textbf{Value} \\
\midrule
\multirow{5}{*}{Data \& model}
    & Timestep $\Delta t$               & $0.4$\,s \\
    & Observation horizon $H_{\text{obs}}$               & $8$ steps ($3.2$\,s) \\
    & Prediction horizon $T$            & $12$ steps ($4.8$\,s) \\
    & Discrete modes $K_z$ / retained $K$ & $6$ / $3$ \\
    & Diffusion steps (train / infer)   & $100$ / $5$ (DDIM) \\
\midrule
\multirow{2}{*}{Baselines}
    & MID-Ens trajectory samples $M$    & $64, 128, 256$ \\
    & MID-Ens GMM components $K$        & $3$ \\
\midrule
\multirow{1}{*}{Calibration}
    & Data split (train / test / cal)   & $70\%$ / $20\%$ / $10\%$ \\
\midrule
\multirow{5}{*}{Controller}
    & MPPI rollouts $N$                     & $1024$ \\
    & Planning horizon $H$          & $4$ steps ($1.6$\,s) \\
    & MPPI temperature $\lambda$        & $1.5$ \\
    & Noise correlation (AR(1)) $\rho$  & $0.8$ \\
    & Per-axis velocity limit           & $\pm0.6$\,m/s \\
\midrule
\multirow{3}{*}{Cost weights}
    & Goal weight $w_{\text{goal}}$     & $50$ \\
    & Wall weight $w_{\text{wall}}$     & $1000$ \\
    & GMM weight $w_{\text{gmm}}$       & $30$ \\
\midrule
\multirow{2}{*}{Chance constraint}
    & Risk budget $\delta$              & $0.1$ \\
    & Softplus temperature $\tau$       & $0.1$ \\
\midrule
\multirow{3}{*}{Geometry}
    & Robot / pedestrian collision radius $r_r/r_h$ & $0.3$\,m / $0.3$\,m \\
    & Wall clearance margin $d_{\text{wall}}$ & $0.25$\,m \\
    & Softplus width $\epsilon$         & $0.05$\,m \\
\bottomrule
\end{tabular}
\end{table}

\subsection{Inference Time Comparison}\label{app:inference_time}

Table~\ref{tab:inference_time_full} compares the per-step inference time of \textbf{Banded+LR} (Ours) against \textbf{MID-Ensemble} at ensemble sizes $M\in\{64,128,256\}$, measured on the $N_p{=}3$ pedestrian scenario (same setting as main-text Table~\ref{tab:inference_time}), averaged over all prediction steps on a single NVIDIA A5000 GPU.

All MID-Ensemble diffusion passes are fully batched into a single GPU call to ensure a fair comparison; the reported times reflect wall-clock cost under optimal GPU parallelism. Diffusion sampling accounts for over $97\%$ of MID-Ensemble prediction time, and its cost increases with the sample budget. Banded+LR requires only a single diffusion pass regardless of the desired uncertainty resolution, with the precision tube head adding $0.60$\,ms and calibration adding $0.08$\,ms.

\begin{table}[ht]
    \centering
    \caption{Inference time breakdown ($N_p{=}3$): MID-Ensemble at $M\in\{64,128,256\}$ samples versus Banded+LR (Ours). All diffusion passes are fully batched on a single GPU. Ours shares the same backbone but avoids repeated sampling, yielding consistently lower total latency.}
    \label{tab:inference_time_full}
    \small
    \begin{tabular}{l|cccc}
        \toprule
        \textbf{Method} & \textbf{Diffusion [ms]} & \textbf{Head/GMM [ms]} & \textbf{Calibration [ms]} & \textbf{Total [ms]} \\
        \midrule
        MID-Ens ($M{=}64$)  & 26.246 & 0.336 & -- & 26.581 \\
        MID-Ens ($M{=}128$) & 47.083 & 0.444 & -- & 47.527 \\
        MID-Ens ($M{=}256$) & 91.222 & 0.558 & -- & 91.779 \\
        \midrule
        Banded+LR (Ours)    & \textbf{12.19} & 0.60 & 0.08 & \textbf{12.87} \\
        \bottomrule
    \end{tabular}
\end{table}

\textbf{Scaling with pedestrian count.}
As the number of pedestrians increases, MID-Ensemble latency grows further because the batch size scales with $N_p$. At $N_p{=}9$ with $M{=}256$, prediction latency reaches approximately $200$\,ms. Pedestrians continue moving during computation in our latency-aware evaluation. If a new forecast is unavailable at a control update, the controller uses interpolated predictions from the previous forecast. Higher latency therefore delays the incorporation of new observations into control. Banded+LR avoids the additional $M$-sample ensemble factor, requiring only one diffusion decode per retained mode to estimate uncertainty.

\begin{figure}[h]
    \centering
    \includegraphics[width=0.55\linewidth]{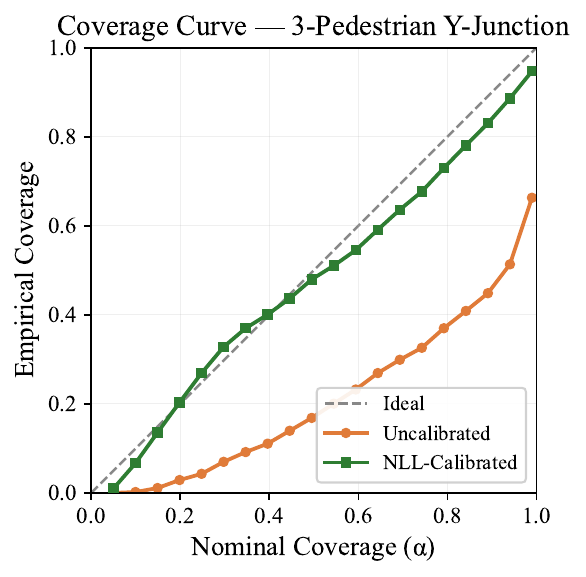}
    \caption{Coverage diagnostic on the 3-pedestrian Y-junction. NLL calibration improves empirical coverage relative to the uncalibrated tubes, but is not coverage-targeted and does not provide formal coverage guarantees.}
    \label{fig:coverage_diagnostic}
\end{figure}
\textbf{Coverage diagnostic.} 
Although our post-hoc calibration is optimized for trajectory-level
negative log-likelihood rather than target coverage, we additionally report empirical coverage
curves as a diagnostic. For each nominal coverage level $\alpha$, we evaluate whether the
ground-truth future lies inside the corresponding marginal Gaussian confidence ellipse induced
by the predicted covariance. No additional coverage-specific scaling or conformal correction is
fit for this plot. As shown in Fig.~\ref{fig:coverage_diagnostic}, the uncalibrated tubes under-cover, while the NLL-calibrated tubes move much closer to the ideal diagonal. This suggests
that likelihood calibration also improves the practical scale of the uncertainty tubes, but the plot
should be interpreted only as a diagnostic rather than a formal coverage guarantee.

\subsection{Maze2D MPPI Sample-Efficiency}\label{app:exp_maze2d}

\textbf{Maze2D Concave environment.}
\begin{table}[ht]
\centering
\caption{Hyperparameters for the Maze2D experiment (Q2). The diffusion prior is shared by Diff-MPPI and Ours;
the Precision Tube Head is exclusive to Ours; the MPPI block applies to
all four methods.}
\label{tab:exp_maze2d_hp}
\small
\begin{tabular}{llr}
\toprule
\multicolumn{3}{l}{\textbf{Diffusion trajectory prior} (Diff-MPPI and Ours)} \\
\midrule
Backbone              & 1D temporal U-Net      & dim $64$, mults $(1,4,8)$ \\
Wall encoder          & 1D vision transformer  & embed $64$, depth $3$ \\
Horizon $H$           &                        & $48$ \\
Diffusion steps (train / infer)  &             & $100$ / $5$ (DDIM) \\
CFG dropout / sampling weight  &               & $0.2$ / $2.0$ \\
Training steps        &                        & $2\!\times\!10^{6}$ \\
EMA decay             &                        & $0.995$ \\
\midrule
\multicolumn{3}{l}{\textbf{Precision Tube Head} (Ours only)} \\
\midrule
Low-rank rank $r$     &                        & $20$ \\
Markov order          &                        & $4$ \\
Trace-CV regulariser $(\lambda, \text{floor})$ & & $(1.0, 0.15)$ \\
\midrule
\multicolumn{3}{l}{\textbf{MPPI controller} (all four methods)} \\
\midrule
Planning horizon $T_{\text{mppi}}$ &           & $15$ \\
Temperature $\lambda$ &                        & $0.169$ \\
Timestep $\Delta t$   &                        & $0.1$ \\
Base noise $\sigma_{\text{base}}$ (vanilla / diff) & & $1.0$ \\
Base noise $\sigma_{\text{base}}$ (log)        & & $0.8$ \\
Sample-count sweep $N$ &                       & $4,8,\ldots,1024$ \\
Arclength lookahead $L_{\text{dist}}$          & & $0.4$ \\
\bottomrule
\end{tabular}
\end{table}
We evaluate on a 2D point-robot navigation task in a $5\times 5$ continuous
maze. Each maze contains $N=7$ axis-aligned rectangular walls of half-extent
$0.5$, of which a subset are arranged in L-shapes. These concave
configurations deliberately create local minima that are difficult for
greedy sampling-based controllers such as MPPI, since the goal-distance
gradient often points directly into an obstacle. The robot has state
$\mathbf x_t=(p_x,p_y)\in[0,5]^{2}$ and action $\mathbf u_t=(v_x,v_y)\in[-1,1]^{2}$, with
single-integrator dynamics $\mathbf x_{t+1}=\mathbf x_{t}+\mathbf u_{t}\,\Delta t$ at $\Delta t=0.1$,
and a body radius of $0.01$ for collision checking.

\textbf{Training dataset.}

The training dataset from PB-diff~\citep{luo2024potentialbaseddiffusionmotion} comprises $3{,}000$ unique mazes with roughly $780$
demonstration trajectories per maze ($\sim$$2.34$M trajectories,
$\sim$$75$M timesteps in total), generated by a feasible-motion planner
over randomised start/goal pairs. Trajectories are loaded with
sliding-window indexing at horizon $H=48$. The same dataset is used to
train the diffusion trajectory prior shared by Diff-MPPI and Ours.

\textbf{Evaluation protocol.}

We evaluate on $100$ held-out mazes with $20$ pre-generated
$(\text{start},\text{goal})$ problems per maze, for a total of $2{,}000$ evaluation problems per configuration. Each diffusion-augmented method samples a single latent style $z$ uniformly and decodes one trajectory per problem.
We do not draw a batch of candidate diffusion paths or report best-of-candidate performance. Each evaluation episode runs for at most $200$ simulation steps with a goal-reach threshold of $0.3$. A trial is
considered as successful if and only if the agent (i)~reaches the goal and
(ii)~remains collision-free along the entire executed trajectory.
Collisions are checked at every executed waypoint and at linearly
interpolated points spaced every $0.05$ units in configuration space,
which avoids missing thin-wall passages.

\textbf{MPPI sample-count sweep.}
To characterise sample efficiency, we evaluate each method at $N\in\{4,8,16,32,64,128,256,512,1024\}$ MPPI rollouts per planning step, re-running all $2{,}000$ problems per $(N,\text{method})$ pair. Success-rate-versus-$N$ curves are reported in the main paper (Fig.~\ref{fig:maze2d_SR_vs_numsamples}).

\textbf{Baselines.}

We compare four sampling-based local planners. All four share the same
controller skeleton (receding-horizon MPPI with single-integrator
dynamics, identical goal cost, and identical collision penalty); they
differ only in how the perturbation distribution is constructed and, for
the diffusion-augmented variants, whether a learned global path is
provided as a tracking reference.
\begin{itemize}[leftmargin=*]
    \item \textbf{Vanilla MPPI}~\cite{williams2017mppi} is the standard
          formulation. Action perturbations are drawn from an isotropic
          Gaussian
          $\delta\mathbf u\sim\mathcal{N}(\mathbf 0,\sigma_{\text{base}}^{2}\mathbf I)$ around
          the rolled-out nominal control sequence. The running cost
          combines a squared-$L_2$ goal-distance term and a hard binary
          obstacle penalty ($10^{7}$ if the rolled state lies inside any
          wall inflated by the robot radius, $0$ otherwise). A
          terminal cost equal to the final-step goal distance times the
          planning horizon is added for long-range guidance.
    \item \textbf{Log-MPPI} retains the cost and dynamics of Vanilla MPPI
          but replaces the noise distribution with a Normal $\times$
          LogNormal mixture
          $\delta \mathbf u\sim\mathcal{N}(\mathbf 0,\sigma_{\text{base}}^{2}\mathbf I)\,\odot\,
          \mathrm{LogN}(\boldsymbol\mu_{\mathrm{LN}},\boldsymbol\sigma_{\mathrm{LN}}^{2})$,
          with element-wise product. The heavy upper tail of the
          LogNormal multiplier produces occasional large perturbations (hence, we use a slightly lower exploration covariance than for Vanilla MPPI)
          that help the controller escape the local minima created by
          L-shaped obstacles, at the cost of slower convergence in open
          regions.
    \item \textbf{Diff-MPPI} introduces a global path prior.
    A diffusion model trained on the demonstration dataset generates a trajectory conditioned on the start, goal, and wall geometry of the current maze. MPPI then tracks this path using an arclength-based lookahead subgoal
          ($L_{\text{dist}}=0.4$), so the running cost penalises
          deviation from a moving subgoal rather than from the global
          goal directly. Sampling is still isotropic Gaussian.
    \item \textbf{Ours} reuses the diffusion prior of Diff-MPPI but additionally conditions a Precision Tube Head on the selected path to produce a per-timestep covariance tube $\boldsymbol\Sigma_{t}$ over actions.
    MPPI samples are drawn from $\delta\mathbf u_{t}\sim\mathcal{N}(\mathbf 0,\boldsymbol\Sigma_{t})$.
          The head
          concentrates samples in the directions the diffusion prior is
          confident about and widens them where the prior is uncertain,
          which is what enables sample-efficient tracking.
\end{itemize}

\textbf{Diffusion trajectory prior.}
For Maze2D, Diff-MPPI and Ours share a multimodal style-conditioned diffusion backbone, implemented as a 1D temporal U-Net with base width $64$ and channel multipliers $(1,4,8)$.
Wall geometry is encoded by a 1D vision transformer with embedding dimension $64$ and depth $3$.
Start and goal are injected via FiLM modulation.
At evaluation time, each diffusion-augmented method samples one $z$ uniformly and decodes one reference trajectory per problem.
We do not enumerate multiple $z$ values, draw a batch of diffusion trajectories, or select the best candidate, so that the diffusion-augmented controllers are evaluated under the same single-plan protocol as the diffusion planners.
Preliminary tests of SCOPE with a standard DDPM backbone without discrete style conditioning yielded similar navigation performance.

We use standard DDPM training with $S=100$ diffusion steps, $\epsilon$-prediction, and classifier-free guidance with concept-drop probability $0.2$ (sampling weight $w_{\text{cfg}}=2.0$).
At inference we use DDIM~\citep{song2022denoisingdiffusionimplicitmodels} with $5$ sampling steps.
The model is trained for $2\times 10^{6}$ steps with batch size $512$, learning rate $2\times 10^{-4}$, and EMA decay $0.995$.
The same trained checkpoint is shared between Diff-MPPI and Ours.

\textbf{Precision Tube Head.}

For Maze2D, the Precision Tube Head predicts a banded-plus-low-rank
precision matrix per action timestep along a conditioning path. It uses
rank $r=20$, hidden width $256$, FiLM conditioning on the diffusion U-Net
features, and Markov order $4$. The diagonal is parameterised in
log-precision space. Training uses AdamW with
learning rate $1\times 10^{-4}$ on a cosine schedule (warm-up $5{,}000$
steps), weight decay $1\times 10^{-4}$, batch size $256$, and $200$
epochs. A trace coefficient-of-variation floor regulariser
($\lambda_{\text{trace-cv}}=1.0$, floor $=0.15$) is applied during
training; in preliminary runs without this regulariser the head collapsed
to nearly isotropic predictions and lost the per-timestep tube structure
that drives the sample-efficiency gains.

\textbf{MPPI controller.}
All four methods share the same MPPI loop; method-specific noise distributions and shared controller settings are listed in Table~\ref{tab:exp_maze2d_hp}. Diff-MPPI and Ours use arclength-based subgoal tracking.

\afterpage{%
\clearpage
\begin{figure}[p]
    \centering
    \includegraphics[
        width=0.82\textwidth,
        height=0.82\textheight,
        keepaspectratio
    ]{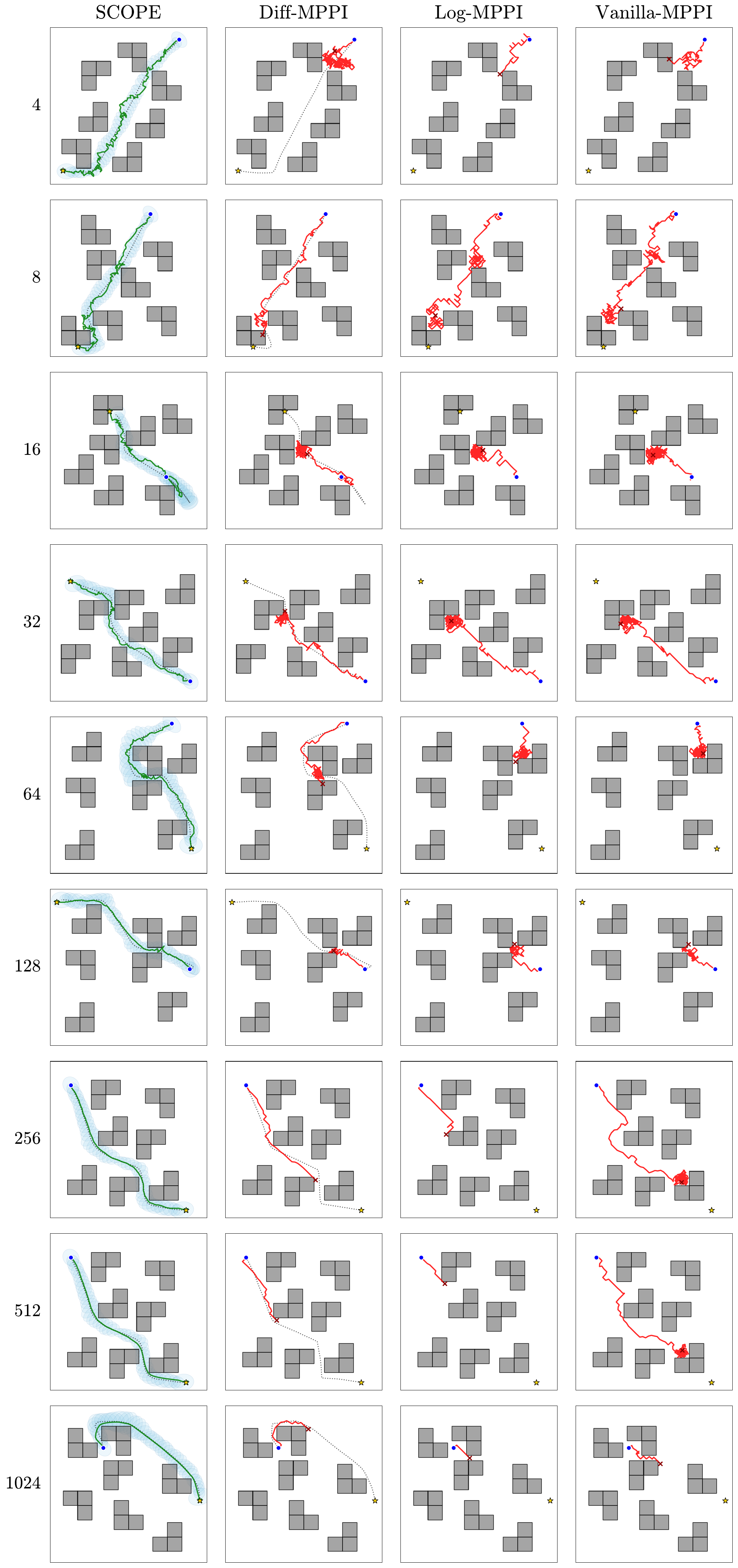}
    \caption{
    Qualitative comparison of MPPI variants on selected challenging Maze2D instances at different rollout budgets
    $N\in\{4,8,16,32,64,128,256,512,1024\}$ (rows).
    Within each row, all methods are evaluated on the same maze and start--goal pair.
    Green trajectories denote collision-free executions, while red trajectories denote collisions ($\times$).
    The blue-shaded region around the SCOPE trajectory visualizes the predicted uncertainty tube.
    The blue dot and star denote the start and goal, respectively.
    For SCOPE and Diff-MPPI, the dashed gray line denotes the shared diffusion-prior trajectory.
    Across these selected cases, SCOPE reaches the goal without collision at all shown rollout budgets,
    whereas the baselines require larger rollout budgets in several cases and may still collide or become trapped in concave obstacle regions.
    }
     \label{fig:maze_methods_comparison}
\end{figure}
\clearpage
}

\subsubsection{Maze2D Navigation Results}\label{app:maze2d_results}

Table~\ref{tab:maze2d} reports the full Maze2D results, including pure diffusion planners.
Diffuser and PB-Diff achieve high goal-reaching rates, but their collision rates remain high.
This shows that diffusion provides a useful global trajectory prior, but does not by itself guarantee local feasibility in held-out maze layouts.
The generated trajectory may reach the goal in a global sense, while still cutting through a wall or violating the exact obstacle constraints along the way.
Thus, goal reaching alone is not sufficient for safe execution in this benchmark.

Adding MPPI substantially improves safety because candidate rollouts are evaluated against the current maze geometry.
Vanilla MPPI and Log-MPPI reduce collision rates, but their goal rates remain lower because short-horizon sampling can get trapped in the concave basins created by L-shaped obstacles.
Diff-MPPI improves goal reaching by using the diffusion trajectory as a global reference, but it still relies on a fixed isotropic proposal covariance.
This creates a single-scale trade-off.
With small covariance, MPPI stays too close to the diffusion reference and cannot repair local errors.
With large covariance, many rollouts near walls collide, and the MPPI update is dominated by a small set of near-boundary survivors.
This can lead to obstacle-skimming executions and explains why Diff-MPPI improves goal rate but still has a higher collision rate than Vanilla MPPI and Log-MPPI.

\begin{table}[t]
    \centering
    \caption{Maze2D navigation results. MPPI-based planners use $N=1024$ rollouts per planning step.}
    \label{tab:maze2d}
    \small
    \renewcommand{\arraystretch}{1.15}
    \setlength{\tabcolsep}{6pt}
    \begin{threeparttable}
    \begin{tabular*}{\linewidth}{@{\extracolsep{\fill}}lcccc@{}}
        \toprule
        \textbf{Planner} & SR$\uparrow$ & CR$\downarrow$ & GR$\uparrow$ & $\overline L$\,[m]$\downarrow$  \\
        \midrule
        Diffuser          & 0.874  & 0.126  & 1.0    & 6.134  \\
        PB-Diff          & 0.887  & 0.113  & 1.0    & 6.042 \\
        \midrule
        Vanilla MPPI & 0.9375 & 0.0125 & 0.9500 & 5.573 {\scriptsize$\pm$ 3.354}  \\
        Log-MPPI & 0.9455 & \textbf{0.0065} & 0.9510 & 5.994 {\scriptsize$\pm$ 4.506}  \\
        Diff-MPPI & 0.9635 & 0.0260 & 0.9875 & 5.328 {\scriptsize$\pm$ 1.601}  \\
        \mymethod{} (Ours) & \textbf{0.9925} & \textbf{0.0065} & \textbf{0.9990} & \textbf{4.996} {\scriptsize$\pm$ 0.763} \\
        \bottomrule
    \end{tabular*}
    \begin{tablenotes}[flushleft]\footnotesize
        \item SR: success rate, CR: collision rate, GR: goal rate, $\overline L$: avg.\ path length. \textbf{Bold}: best.
    \end{tablenotes}
    \end{threeparttable}
\end{table}

SCOPE addresses this issue by using the predicted tube as a trajectory-dependent proposal covariance.
The proposal remains broader where the diffusion prior is flexible, and becomes tighter in constrained regions where feasible perturbations are limited.
As a result, SCOPE combines the global guidance of diffusion with local, constraint-aware refinement from MPPI.
Compared with Diff-MPPI, which uses the same diffusion prior but a fixed covariance, SCOPE achieves higher success rate, lower collision rate, and shorter paths.
This confirms that the improvement comes from making the diffusion-to-MPPI interface uncertainty-aware, rather than from the diffusion prior alone.

Figure~\ref{fig:maze_methods_comparison} visualizes selected challenging trajectories produced by each MPPI variant across rollout
budgets $N\in\{4,8,16,32,64,128,256,512,1024\}$.
For each rollout budget, all methods are evaluated on the same maze and start--goal pair.
The shaded tube contracts near obstacles and expands in open corridors, confirming that the learned covariance concentrates rollouts where feasibility constraints are tight and permits broader exploration elsewhere.
By contrast, Diff-MPPI frequently collides or stalls at low budgets because its fixed isotropic covariance wastes rollouts on wall-penetrating samples; even at $N{=}1024$ the trajectory still exhibits obstacle-skimming behaviour.
Log-MPPI and Vanilla MPPI lack global guidance entirely, causing the planner to become trapped in concave regions across most budgets.
These qualitative patterns are consistent with the quantitative gains in Table~\ref{tab:maze2d}: SCOPE's adaptive proposal turns a small rollout budget into a large effective one, yielding the highest success rate and lowest collision rate across all settings.

\subsection{Y-Corridor Closed-Loop Navigation}\label{app:exp_ycorridor}

This experiment evaluates the same MID checkpoints and Precision Tube Head
trained for Q1 inside an MPPI navigation controller. The model checkpoints,
training procedure, and per-density configuration are identical to Q1. The
only difference is the evaluation environment, in which pedestrians are
simulated online by ORCA rather than replayed from the dataset, and the
predictor's GMM output is consumed by an MPPI controller that drives a
holonomic robot through the junction.

\textbf{Closed-loop simulator.}

The simulator instantiates the same Y-corridor used to generate the Q1
dataset, with $N_p\in\{3,6,9,12\}$ ORCA pedestrians. The robot's task is
to navigate from a randomised start to a randomised goal across the
junction while avoiding both walls and pedestrians. The simulation runs at
$\Delta t=0.4$\,s. The robot is holonomic with each component of its velocity
command $\mathbf u_t=(v_x, v_y)$ clipped to
$[-0.6, 0.6]$\,m/s. Robot and pedestrian collision radii are
both $0.3$\,m.

\textbf{Episodes and metrics.}

For each pedestrian count we run $100$ episodes and report success rate
(reaching the goal without collision or timeout), path length, and time to goal
(the latter two over successful episodes only). 

\textbf{Baselines.}

All forecasters expose the same GMM interface ($H_{\text{obs}}=8$ history, $T=12$ future steps, per-mode $2\times 2$ covariances). The MPPI controller consumes every baseline identically through the chance-constraint cost (Sec.~\ref{app:mppi_cost}).
\begin{itemize}[leftmargin=*]
    \item \textbf{NF} (no future) assumes pedestrians remain static at their current position.
    \item \textbf{LKF} is the constant-velocity Kalman filter from Q1. It outputs a single Gaussian.
    \item \textbf{MID-UM} (unimodal) samples a single trajectory from the MID prior.
    \item \textbf{MID-MM} (multimodal) emits top-$K{=}3$ mode mean trajectories.
    \item \textbf{MID-Ens-$M$} extends the Q1 MID-Ens baseline with $M\in\{64,128,256\}$ samples across top-$K{=}3$ modes.
    \item \textbf{Ours-noCal} is the proposed forecaster with calibration disabled, isolating the calibration effect.
    \item \textbf{Ours} is the proposed forecaster with per-horizon $\tau_t$ calibration applied unchanged from Q1.
\end{itemize}

\textbf{MPPI controller.}
The controller uses the cost formulation from Sec.~\ref{app:mppi_cost} (Eqs.~\eqref{eq:goal_cost},~\eqref{eq:wall_cost}, and~\eqref{eq:gmm_cost}), with all pedestrian interaction handled exclusively through the GMM chance-constraint cost. Controller hyperparameters are listed in Table~\ref{tab:exp_pth_hp}. All diffusion calls for both MID-Ens-$M$ and Ours are fully batched into a single GPU diffusion decode path per planning step.

\label{tab:exp_y_hp}

\subsection{Direct NLL Training of the Precision Head}
\label{app:direct_nll_ph}

\begin{figure}[t]
    \centering
    \includegraphics[width=\linewidth]{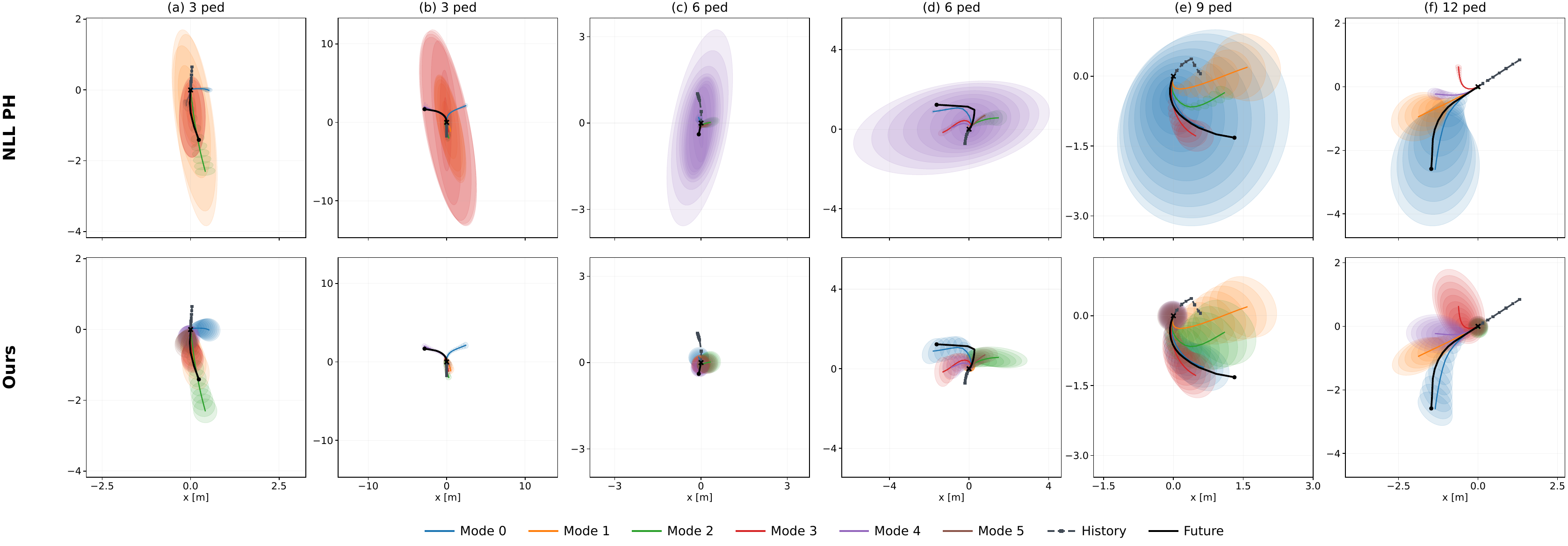}
    \caption{Qualitative comparison of uncertainty tubes from Direct NLL PH (top) and SCOPE (bottom). Columns show selected examples with 3, 6, 9, and 12 pedestrians. Colors identify predicted modes. Shaded ellipses visualize uncertainty along each mode. Dashed gray lines show observed histories, and solid black lines show ground-truth futures. Each column uses matching axis limits across rows. Several Direct NLL PH tubes extend far beyond their nominal trajectories, whereas SCOPE produces more localized uncertainty in these cases.}
    \label{fig:direct_nll_ph_tubes}
\end{figure}

We compare SCOPE with a precision head (PH) trained directly by minimizing negative log-likelihood (NLL), denoted as Direct NLL PH.
Direct NLL PH uses the same precision-head architecture as SCOPE, with the diffusion backbone and nominal trajectories held fixed during training.
Table~\ref{tab:direct_nll_ph_sr} reports navigation success rates (SR) across crowd densities.
Both methods perform equally in the sparsest setting.
As crowd density increases, Direct NLL PH exhibits a larger decline in success rate, and its gap to SCOPE widens.

\begin{table}[htbp]
    \centering
    \caption{Navigation success rate across pedestrian densities. Bold indicates the best result for each density.}
    \label{tab:direct_nll_ph_sr}
    \small
    \begin{tabular}{lcccc}
        \toprule
        & \multicolumn{4}{c}{SR [\%] $\uparrow$} \\
        Method & 3 pedestrians & 6 pedestrians & 9 pedestrians & 12 pedestrians \\
        \midrule
        Direct NLL PH & \textbf{100} & 91 & 84 & 81 \\
        SCOPE (Ours) & \textbf{100} & \textbf{96} & \textbf{93} & \textbf{92} \\
        \bottomrule
    \end{tabular}
\end{table}

Direct NLL PH is sensitive to complex pedestrian interactions in constrained spaces.
Near the junction center, pedestrian motion is influenced by the potential movement of surrounding pedestrians.
Figure~\ref{fig:direct_nll_ph_tubes} shows selected examples in which Direct NLL PH produces broad tubes that extend far beyond the corresponding nominal trajectories.
This expansion is particularly pronounced in panels (a), (e), and (f).
SCOPE produces more localized tubes in these cases, with uncertainty that varies across modes and along the predicted trajectories.

With the nominal trajectories held fixed, direct NLL training can absorb trajectory prediction errors into the estimated covariance. Expanding the covariance along directions with large residuals reduces the Mahalanobis penalty, at the cost of a larger log-determinant term. Thus, wider prediction tubes may reflect errors in the nominal trajectory rather than greater within-mode variability.

Excessively broad tubes can impair downstream control through the collision-risk cost.
For a candidate robot position outside the collision radius of a predicted pedestrian mean, increasing uncertainty along the separation direction raises the union bound risk $P_{\text{risk}}$ used by our controller.
This can penalize otherwise usable routes and discourage progress.
The effect can become more restrictive in dense crowds, where risk contributions from multiple pedestrians accumulate.
The widening SR gap in Table~\ref{tab:direct_nll_ph_sr} is consistent with this mechanism.

\subsection{Real-Robot Experiments}
\label{app:exp_real_robot}
\begin{figure}[htbp]
    \centering
    \includegraphics[width=\linewidth]{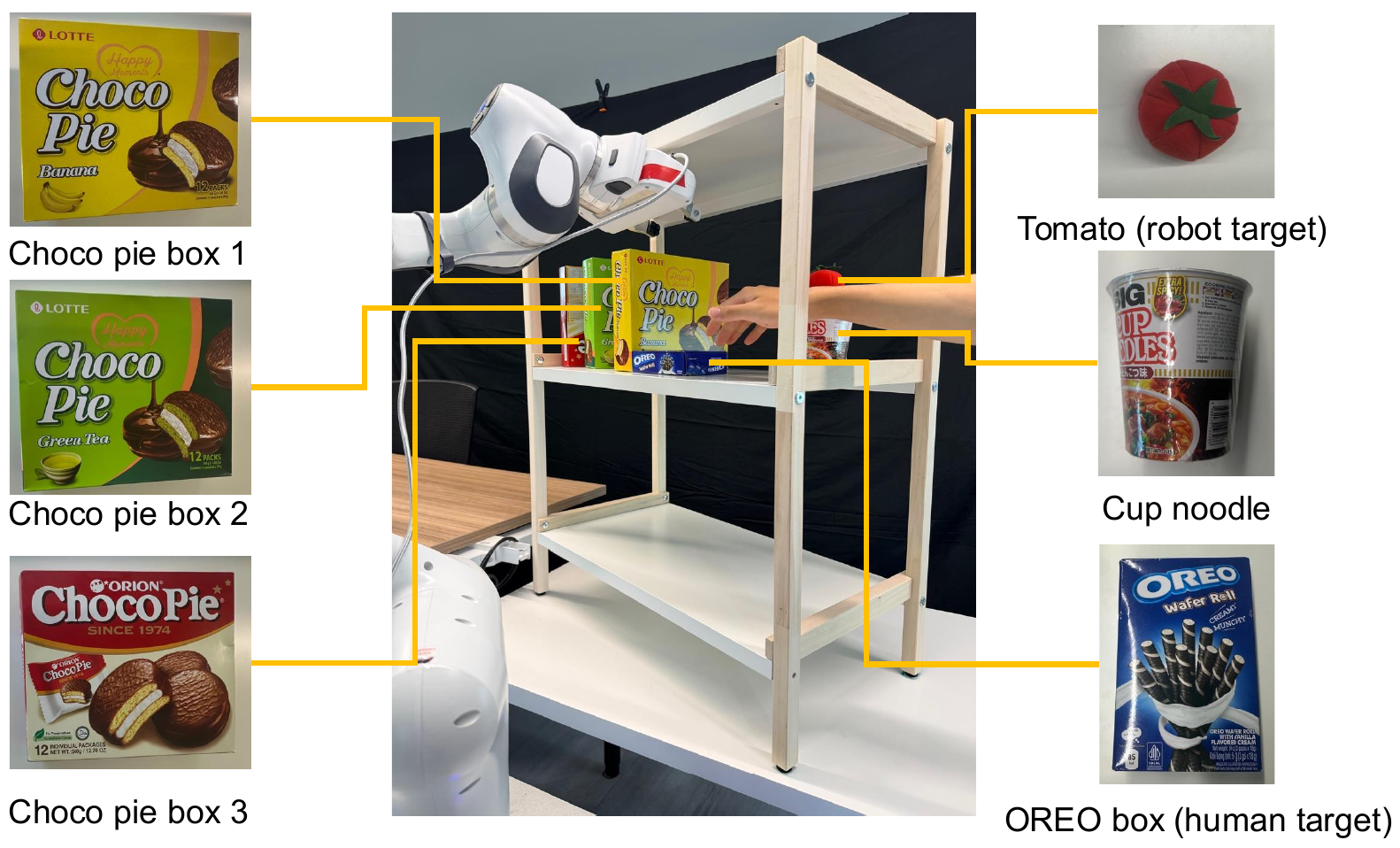}
    \caption{
    Real-robot shelf setup for pick-and-place task.
    The Franka arm operates in a constrained shelf workspace.
    In this example, the OREO box is the human target and the tomato is the robot target, while the remaining objects serve as static obstacles for collision-aware MPPI control.
    }
    \label{fig:app_real_setting}
\end{figure}
\subsubsection{Environment and Hardware}
We conduct the real-robot experiments on a 7-DOF Franka Emika Panda arm equipped with a Franka Hand gripper.
The robot operates in a tabletop shelf setup, where the target object (tomato) is placed inside the shelf and must be transferred to the upper shelf.
The shelf structure and nearby tabletop objects create a constrained, non-convex workspace.

Static obstacles include the shelf, three movable box obstacles placed near the target object, and a cup noodle under the target object (fixed).
We represent these obstacles using prebuilt cuboid primitives and their associated collision model.
To align this model with the physical scene, we use a wrist-mounted RealSense camera to collect point clouds and calibrate the poses of the cuboid primitives with respect to the robot frame.
When the obstacle layout changes, the corresponding primitive pose is updated before execution.

For the dynamic setting, we use two low-cost RGB webcams to observe the human hand.
The controller runs on a workstation equipped with an NVIDIA RTX 5090 GPU, while the robot is commanded through the Franka control stack.

\subsubsection{Task Setup and Evaluation Protocol}

\begin{figure}[t]
    \centering
    \includegraphics[width=\linewidth]{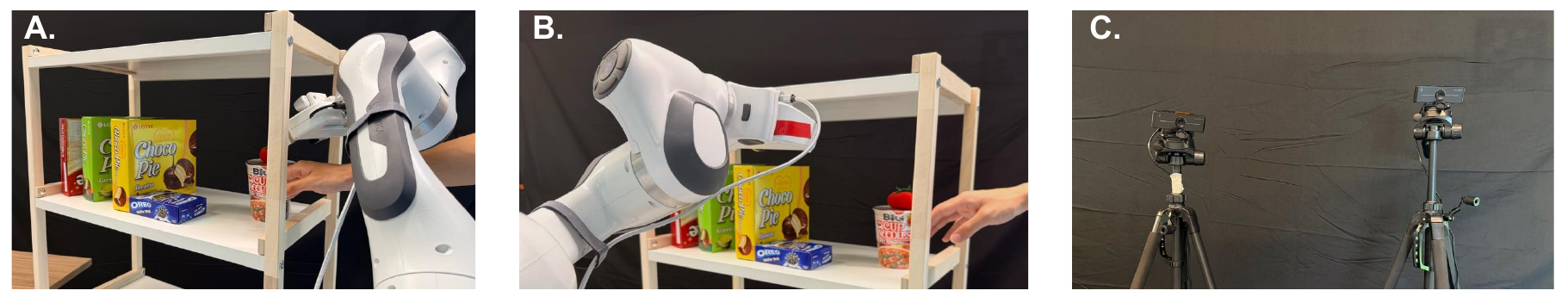}
\caption{
Real-robot setup and representative approach modes.
(A) Front approach, where the arm reaches the target through a wider opening.
(B) Side approach, where the arm reaches the target with tighter clearance.
(C) Low-cost RGB webcam setup used for hand tracking in the dynamic experiments.
}
    \label{fig:app_real_robot_modes}
\end{figure}

The task is to reach into the shelf, grasp the target object (tomato), and place it on the upper shelf.
Across trials, the box obstacles (choco pie boxes) are repositioned to vary the local geometry around the grasp.
The diffusion model produces two common approach modes, shown in Fig.~\ref{fig:app_real_robot_modes}: a front approach and a side approach with a tighter clearance profile.

We evaluate two conditions.
\textbf{Static} uses stationary obstacles and evaluates policy-side tube-informed MPPI control.
\textbf{Dynamic} adds a human hand reaching into the workspace to pick the Oreo box and evaluates whether the predicted hand tube improves avoidance.
Each method is evaluated over 30 trials per condition.
We report success rate (SR), collision rate (CR), and timeout rate (TO).
A timeout is a collision-free trial that fails to complete the task within 120\,s.
In the dynamic setting, we separately report collisions with the environment and with the human hand; a single trial may incur both types.

\subsubsection{Controller and System Implementation}

\paragraph{Static scene calibration.}
The shelf and tabletop obstacles are modeled as cuboid primitives.
A prebuilt collision representation is calibrated to the real scene using point clouds from the wrist-mounted RealSense camera.
This calibration provides the SE(3) poses of the shelf and object primitives used by the controller.
The calibrated primitives are then kept fixed during a trial and queried for collision checking during MPPI rollouts.

\paragraph{Diffusion and \mymethod{}.}
The real-robot diffusion model follows the same diffusion-prior and precision-head design used in the Maze2D experiments, but is adapted to Franka joint-space trajectories.
Each trajectory is represented in an 8D space consisting of the 7 robot joint coordinates and one gripper coordinate.
The conditioning input is changed from the 2D maze layout to the calibrated SE(3) poses of the cuboid scene primitives.
Given the current robot state and scene geometry, the diffusion model generates candidate joint-space trajectories, and the precision tube head predicts a time-varying uncertainty tube around each trajectory.

\paragraph{Mode selection.}
In the real-robot experiments, the diffusion prior produces two approach modes
($K=2$). Equivalently, let $\mathcal{K}=\{1,2\}$ denote the mode set.
For each mode $k\in\mathcal{K}$, we compute
\[
w_k = \min_i \lambda_{\min}(\Sigma_{k,i}),
\]
where $\Sigma_{k,i}$ is the marginal covariance at waypoint $i$ in normalized trajectory space. The selected mode is
\[
k^\star = \arg\max_{k\in\mathcal{K}} w_k .
\]
This biases execution toward trajectories that remain flexible at their tightest point, rather than choosing
an approach that becomes highly constrained near the shelf.

\paragraph{Tube-informed MPPI controller.}
The selected diffusion trajectory is converted into a smooth joint-space reference. The diffusion trajectory acts as a soft guide for MPPI rather than a committed execution path.
The MPPI controller then refines this reference online in a receding-horizon loop.
Rather than tracking the entire trajectory as a fixed goal, the controller follows a local reference along the diffusion trajectory.
The precision tube is used to shape the MPPI sampling distribution over time: regions with larger predicted flexibility allow broader exploration, while narrow and high-precision regions use more conservative perturbations.
This lets the controller refine the diffusion prior while respecting the calibrated shelf and obstacle geometry.

For collision avoidance, the controller evaluates sampled rollouts against the calibrated static primitives.
We use a constraint-aware rollout score similar in spirit to COSMIK-MPPI~\citep{gursoy2026cosmikmppiscalingconstrainedmodel}, where unsafe rollouts are strongly downweighted instead of relying only on large additive collision penalties.
This is important in the shelf task, where many sampled motions can enter collision in tight non-convex regions.

\begin{figure}[t]
    \centering
    \includegraphics[width=\linewidth]{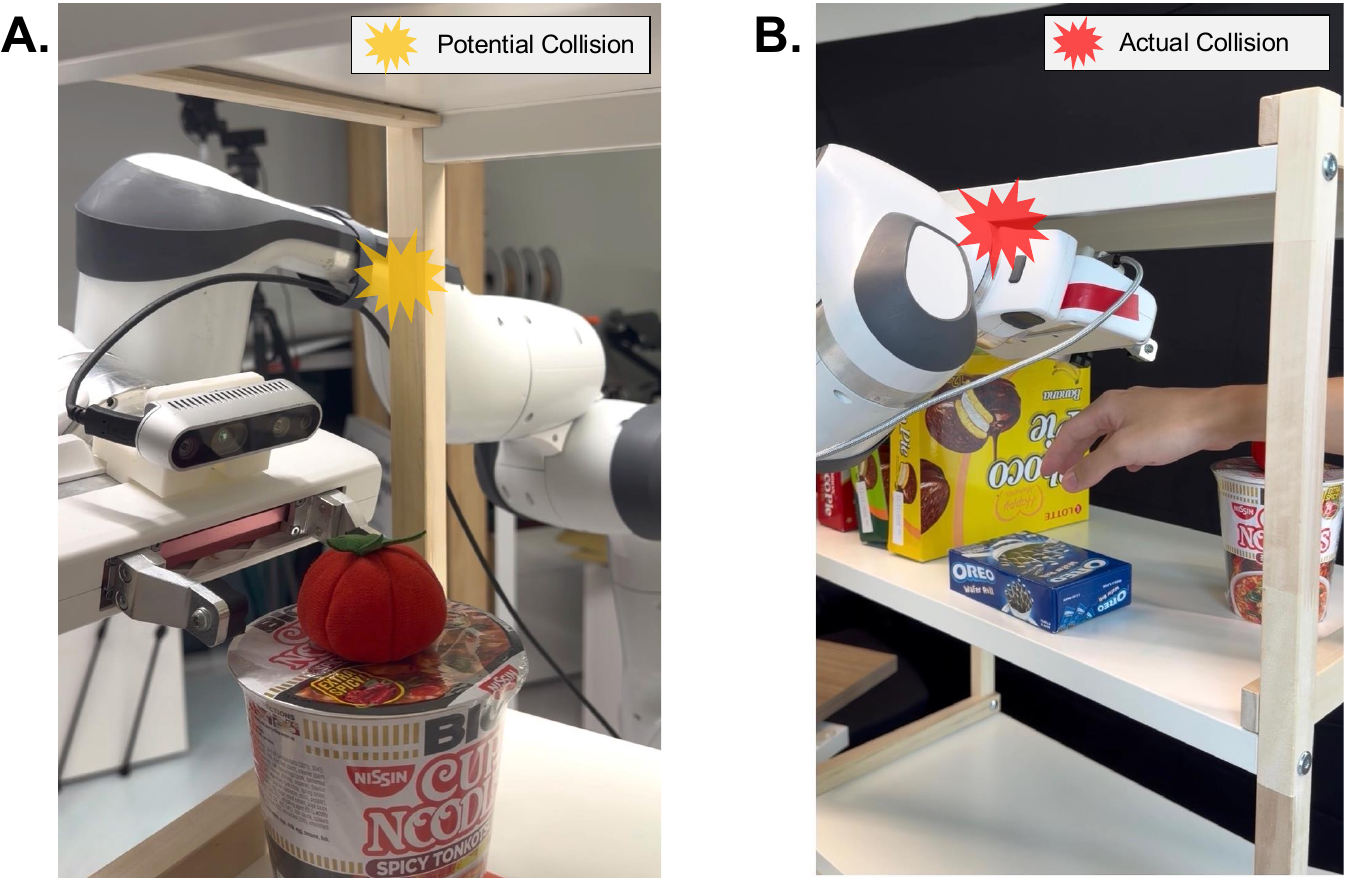}
\caption{
Representative real-robot failure cases.
\textbf{A.} Static Diff-MPPI failure.
Without SCOPE, the planner lacks clearance- and flexibility-aware mode selection and may choose the tighter side approach.
Near the object, a potential shelf collision blocks further progress; limited-horizon MPPI cannot retreat and reorient, so the trial times out (TO).
\textbf{B.} Dynamic failure of Ours (no WM), where SCOPE guides the robot policy but is not used as a human-motion world model.
Without hand prediction, the controller reacts only after the hand enters the shelf workspace, causing a late upward evasive motion that collides with the shelf.
}
    \label{fig:real_robot_case_study}
\end{figure}

\paragraph{Human hand tracking and prediction.}
In the dynamic setting, we restrict human modeling to the hand trajectory.
This is sufficient for the tabletop interaction considered here, where the hand is the dominant moving obstacle near the robot, and avoids requiring task-specific whole-body motion data.
We estimate the hand trajectory using a two-RGB-camera markerless hand-tracking pipeline,
following the setup of COSMIK-MPPI~\citep{gursoy2026cosmikmppiscalingconstrainedmodel}. The resulting hand observations are streamed to the previously described diffusion backbone to predict future hand trajectories.
The predicted hand trajectory is augmented with \mymethod{} and converted into time-varying collision primitives.
The MPPI controller consumes these primitives together with the static scene model, allowing it to yield before the hand reaches the robot's nominal path.

\paragraph{Execution.}
At the beginning of each trial, the scene primitives are calibrated and the current robot state is measured.
The diffusion model generates candidate joint-space trajectories, the precision tube head estimates their uncertainty, and the controller selects one mode for closed-loop execution.
The selected trajectory is not executed open-loop and its waypoints are not imposed as hard constraints.
Instead, it defines a soft state-space reference for MPPI.

The MPPI controller runs at 50 Hz.
At each control tick, the controller projects the measured robot state onto the selected trajectory using a forward-progress index and chooses a lookahead reference ahead of the current progress.
This carrot-like reference is used as one term in the rollout cost, together with collision, joint-limit, and control costs.
Therefore, MPPI can choose rollouts that deviate from the diffusion reference when doing so gives a lower total cost or avoids a collision.
The precision tube further shapes the rollout covariance along the trajectory, allowing broader correction in flexible regions and more conservative exploration near tight passages.

\subsubsection{Real-Robot Case Studies}
\label{app:real_robot_cases}

Fig.~\ref{fig:real_robot_case_study} shows representative failurcases that explain the main error modes in the real-robot experiments.

In the static setting, Diff-MPPI can fail because it uses the diffusion prior
without SCOPE-based clearance and flexibility awareness. In the example in
Fig.~\ref{fig:real_robot_case_study}A, the planner selects the tighter side
approach. This mode can bring the gripper close to the object, but leaves little
room for correction near the shelf. Once a potential collision prevents further
forward motion, the short-horizon MPPI controller may not discover the longer
recovery behavior needed to back out, reorient the end-effector, and retry the
grasp. The robot therefore remains stuck near the object until timeout.

In the dynamic setting, Ours (no WM) still uses SCOPE as a policy-side prior for
the robot motion, but does not use SCOPE as a world model to predict the human
hand. As shown in Fig.~\ref{fig:real_robot_case_study}B, the controller reacts
only after the hand has already entered the shelf workspace. At that point, the
arm is already deep inside the constrained shelf, so the late evasive motion can
lift the arm into the shelf and cause a collision. With the predicted hand tube,
the full method begins yielding earlier, before the robot reaches such a
low-flexibility configuration.

\end{document}